\documentclass{bmvc2k}

\title{On the Role of Individual Differences in Current Approaches to Computational Image Aesthetics}

\addauthor{Li-Wei Chen}{li-wei.chen@kuleuven.be}{1,2}
\addauthor{Ombretta Strafforello}{ombretta.strafforello@kuleuven.be}{1}
\addauthor{Anne-Sofie Maerten}{annesofie.maerten@kuleuven.be}{1}
\addauthor{Tinne Tuytelaars}{Tinne.Tuytelaars@esat.kuleuven.be}{2}
\addauthor{Johan Wagemans}{johan.wagemans@kuleuven.be}{1}

\addinstitution{
Brain and Cognition, KU Leuven, Belgium
}
\addinstitution{
Department of Electrical Engineering (ESAT), KU Leuven, Belgium, KU Leuven, Belgium
}

\runninghead{Chen \etal}{On the Role of Individual Differences in IAA}

\def\ie{\emph{i.e}\bmvaOneDot}
\def\eg{\emph{e.g}\bmvaOneDot}

\def\etal{\emph{et al}\bmvaOneDot}

\begin{document}

\maketitle

\begin{abstract}
Image aesthetic assessment (IAA) evaluates image aesthetics, a task complicated by image diversity and user subjectivity. Current approaches address this in two stages: Generic IAA (GIAA) models estimate mean aesthetic scores, while Personal IAA (PIAA) models adapt GIAA using transfer learning to incorporate user subjectivity. 
However, a theoretical understanding of transfer learning between GIAA and PIAA, particularly concerning the impact of group composition, group size, aesthetic differences between groups and individuals, and demographic correlations, is lacking. 
This work establishes a theoretical foundation for IAA, proposing a unified model that encodes individual characteristics in a distributional format for both individual and group assessments. We show that transferring from GIAA to PIAA involves extrapolation, while the reverse involves interpolation, which is generally more effective for machine learning. 
Extensive experiments with varying group compositions, including sub-sampling by group size and disjoint demographics, reveal substantial performance variation even for GIAA, challenging the assumption that averaging scores eliminates individual subjectivity. Score-distribution analysis using Earth Mover’s Distance (EMD) and the Gini index identifies education, photography experience, and art experience as key factors in aesthetic differences, with greater subjectivity in artworks than in photographs. Code is available at \url{https://github.com/lwchen6309/aesthetics_transfer_learning}.
\end{abstract}
\section{Introduction}\label{sec:intro}
Assessing the aesthetics of images, known as Image Aesthetics Assessment (IAA), is a challenging task due to the inherent complexity of image diversity and individual subjectivity~\cite{talebi2018nima, yang2022personalized, ke2021musiq, yang2023multi, zhu2022personalized, zhu2020personalized, strafforello2024backflip}. IAA has become significant due to the success of image generation~\cite{rombach2022high,podell2023sdxl,ramesh2022hierarchical,saharia2022photorealistic}, which has amplified the need to adjust images according to personal aesthetics~\cite{goetschalckx2019ganalyze}.
Extensive research~\cite{van2021cross, kossmann2023composition, van2023order, wagemans2012century, damiano2023role} has explored how image aesthetics correlate with various image attributes, \eg, spatial composition~\cite{van2021cross,kossmann2023composition,van2023order}, figure-ground organization~\cite{wagemans2012century}, and symmetry~\cite{damiano2023role}. 
Image aesthetics are also influenced by image type, with artworks generally considered more subjective than photographs~\cite{vessel2018stronger}. Additionally, individual differences in perception play a role in image aesthetic~\cite{van2021individual, de2014individual, samaey2020individual}, contributing to individual subjectivity that correlates with demographic factors. This adds further challenges for modeling IAA.

% The existing work taklce Introduce GIAA and PIAA
Current IAA approaches tackle image diversity and individual subjectivity separately in two stages. %, as depicted in Fig.~\ref{fig:giaa_piaa}. 
First, Generic IAA (GIAA) models~\cite{talebi2018nima,yang2022personalized,ke2021musiq} estimate averaged user aesthetic scores or score distributions across a broad range of images, aiming to capture a mean score without individual subjectivity. Subsequently, Personal IAA (PIAA) models~\cite{yang2022personalized,yang2023multi,zhu2022personalized,zhu2020personalized,li2020personality,zhu2021learning,li2022transductive,yan2024hybrid,shi2024personalized} adapt these generic models, fine-tuning them with a small amount of data (\ie few-shot learning), with or without incorporating personal traits to handle the subjectivity. This process represents a form of transfer learning~\cite{zhuang2020comprehensive} from GIAA to PIAA, even though it is not explicitly defined as such in the existing PIAA literature.
% ~\cite{yang2022personalized,yang2023multi,zhu2022personalized,zhu2020personalized,li2020personality,zhu2021learning,li2022transductive,yan2024hybrid,shi2024personalized}.

% \begin{figure}
%     \centering
%     \includegraphics[width=.5\linewidth]{figures/scheme.pdf}
%     \caption{Illustration of the existing IAA approach: The model learns \textcolor{red}{GIAA} to predict group mean aesthetics, associated with averaged traits, and is then fine-tuned on \textcolor{violet}{PIAA} to address individual subjectivity.}
%     \vspace{-0.5cm}
%     \label{fig:giaa_piaa}
% \end{figure}

% Problem:
However, the existing approaches present several limitations.
\textit{1)} Existing PIAA approaches
% ~\cite{yang2022personalized,yang2023multi,zhu2022personalized,zhu2020personalized,li2020personality,zhu2021learning,li2022transductive,yan2024hybrid,shi2024personalized} 
make it difficult to analyze aesthetic differences between groups and individuals, as well as their correlation with demographic factors, since these differences can only be inferred through parameter shifts observed during fine-tuning of PIAA. 
\textit{2)} Although existing GIAA approaches~\cite{talebi2018nima,yang2022personalized,ke2021musiq} assume that individual subjectivity can be minimized by averaging scores, the bias caused by demography may persist within group averages. Furthermore, these methods often oversimplify group composition by overlooking variations in demographic factors and group size, which can introduce bias and affect the fine-tuning of PIAA. For example, a GIAA dataset with a group size of 2 is clearly closer to personal aesthetics (PIAA) than a group of size 100, resulting in better transfer learning. This challenge remains largely unexplored.
\textit{3)} Furthermore, the existing GIAA~\cite{talebi2018nima,yang2022personalized,ke2021musiq} and PIAA~\cite{yang2022personalized,yang2023multi,zhu2022personalized,zhu2020personalized,li2020personality,zhu2021learning,li2022transductive,yan2024hybrid,shi2024personalized} approaches do not adequately address generalization to unseen users, \ie zero-shot learning. Given the high subjectivity of image aesthetics, it is important to investigate model generalization on unseen test users and how it correlates with demographic differences between training and test users.

\begin{figure}[h]
    \centering
    \includegraphics[width=0.9\linewidth]{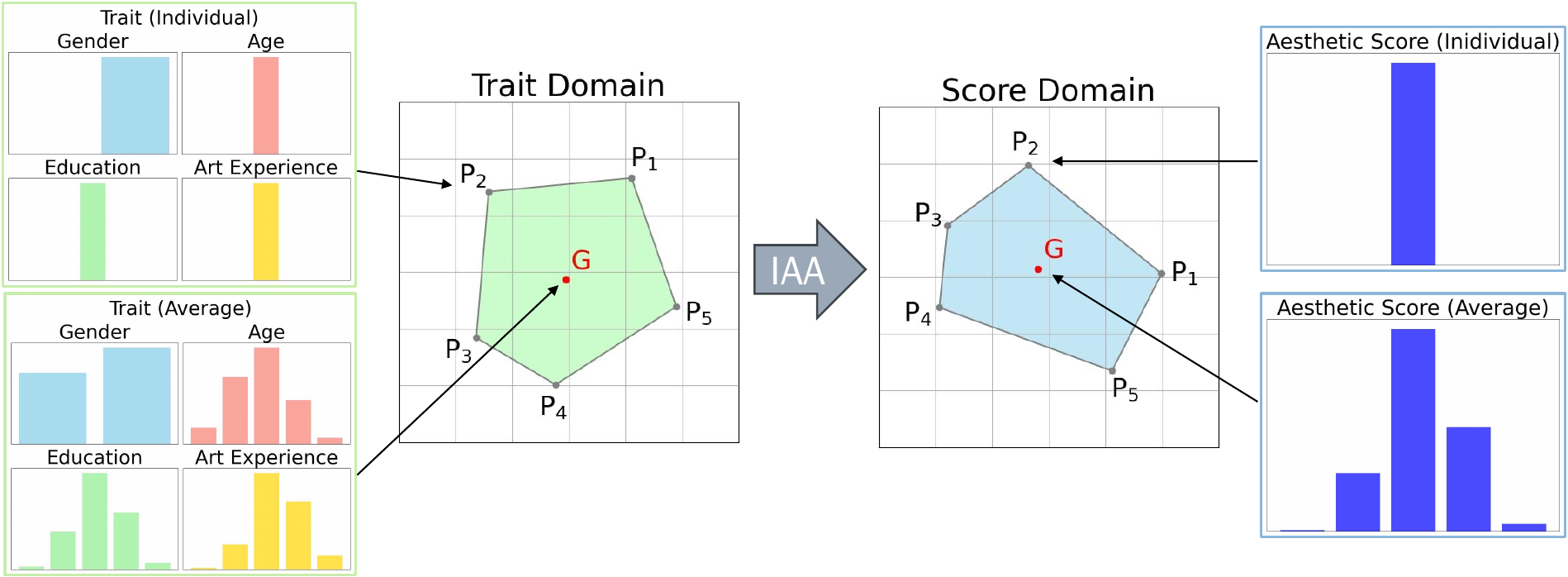}
    \caption{
    The geometry of IAA. For a given input image, the input trait and output score of the IAA are represented by two convex hulls.
    Left: The input convex hull in the space of trait distribution, \eg, age, gender, and education; Right: The output convex hull in the space of aesthetic scores distribution.
    The averaged personal traits distribution of different subsets of individuals all lie within the convex hull formed by the individual traits (\textit{Trait Domain}). 
    These traits can be provided as input to the IAA model, in addition to the input image.
    Similarly, the average aesthetic score distribution given by a group of individuals all lie within the convex hull formed by the aesthetic scores given by single individuals (\textit{Score Domain}).  For both convex hulls, each point $P_i$ represents an individual data point, where $i = 1, 2, 3, \dots$, and $G$ represents the averaged data.
}
    \label{fig:convex_hull}
    % \vspace{-0.6cm}
\end{figure}

To address these issues, we propose a novel IAA approach by encoding personal traits in a distributional format that accounts for both individual and group characteristics. %We input the distributional trait encoding to a single model that performs both GIAA and PIAA.
Our method is capable of inferencing both GIAA and PIAA with a single model by receiving the corresponding trait distribution as input. 
This approach reveals the geometry of the IAA domain, where the input space (personal traits) and output space (aesthetic scores) form distinct convex hulls based on personal data for given images, as depicted in Figure~\ref{fig:convex_hull}. We refer to these convex hulls as the trait convex hull and the score convex hull, respectively. In this context, GIAA maps the average trait distribution located at the inner regime of the trait convex hull to the average score distribution located at the inner regime of the score convex hull. 
In contrast, PIAA maps each vertex of the trait convex hull to corresponding points in the score convex hull. 
Based on this insight, we claim that transfer learning from GIAA to PIAA represents an extrapolation within the characteristic domain, whereas the reverse direction constitutes an interpolation—a generally more effective approach for machine learning models. 
% If this holds, we would expect PIAA models to perform well on GIAA data without GIAA pre-training; a hypothesis that is further confirmed by our experimental results.
If this holds, we would expect PIAA models to perform well on GIAA data without any GIAA pre-training. This hypothesis is further supported by our experimental results.
We extend the GIAA and PIAA baseline models by conditioning them on a distributional trait encoding and demonstrate that direct training on PIAA data achieves performance comparable to the GIAA baseline, whereas training on GIAA data underperforms relative to the PIAA baseline. 
% Then, we modify PIAA training by omitting few-shot sampling and instead use the full training data to ensure a fair comparison between GIAA and PIAA on the same images and users. 
Additionally, we demonstrate that models trained directly on PIAA data match the performance of the GIAA baseline and even exceed the PIAA baseline. 
These results validate our proposed theory regarding interpolation and extrapolation, offering a more precise analysis of the aesthetic differences between groups and individuals.

% Then, we demonstrate the impact of the group's composition in the training set on GIAA models, confirming that averaging individual scores does not necessarily eliminate individual subjectivity.
Then, we demonstrate the transfer learning from GIAA can be significantly affected by group's composition.
We introduce \textit{sGIAA}, a data augmentation method that sub-samples GIAA data, increasing diversity in both demography and group size. By sampling between 2 and the maximum number of users per image, we show that this approach maintains GIAA performance while significantly improving zero-shot PIAA performance by at most 44\%. This improvement arises from increased demographic variation and from the closer training domain to individual users with smaller group sizes. 
% While this approach has negligible effects on GIAA, the significant increase in zero-shot PIAA performance impacts subsequent PIAA fine-tuning. 
% This observation highlights the role of group composition. 

Lastly, we investigate the model 's ability to generalize to unseen users by dividing training and test users into distinct groups, according to demography such as gender and education level. 
We observe a significant performance decrease of up to 80.6\% in this setup compared to seen users, with a larger decrement on the artwork dataset than on the photo dataset.
Our findings reveal that, for both photos and artworks, education and experience with photography and art are the primary factors driving distinct aesthetic judgments. 
% Additionally, we observe a greater variation in model performance on the artwork dataset compared to the photo dataset, which indicates higher individual subjectivity for artworks compared to photographs, which aligns with existing work~\cite{vessel2018stronger}.
% Additionally, we find that metrics of the score distribution, such as Earth Mover's Distance (EMD)~\cite{rubner2000earth}, is an effective index for estimating model performance on unseen users, while the Gini index~\cite{lerman1984note} is an effective index for assessing the significance of demographics to aesthetics.
% We begin by analyzing the score distribution within each demographic category (\eg,, male and female) and calculate the Gini index~\cite{lerman1984note} across these groups to evaluate the effectiveness of demographic-based separation in score distribution. 
% We then evaluate the proposed IAA models on users with distinct demographic groups, calculating the Earth Mover's Distance (EMD)~\cite{rubner2000earth} between their score distributions. Consistent with the Gini index trends, the largest variations in EMD are observed for education, photo experience, and art experience.
This further emphasizes the challenge of PIAA when generalizing to unseen users with varying demographic profiles.
To our knowledge, ours is the first model enabling both GIAA and PIAA. It matches the performance of the GIAA baseline and surpasses state-of-the-art PIAA models that require GIAA pre-training. Furthermore, it is the first theoretical framework to address aesthetic differences between groups and individuals, account for diverse demographic factors and group size. Our work is the first to comprehensively investigate the model's generalization, \ie, its zero-shot performance on unseen users.
\section{Related Work}\label{sec:relate}
\subsection{PIAA models}\label{sec:relate:piaa}
The existing PIAA approaches~\cite{zhu2022personalized,zhu2023personalized,li2020personality,shi2024personalized,yun2024scaling} adapt pre-trained GIAA models through fine-tuning. A frequently chosen model is NIMA~\cite{talebi2018nima}, which predicts score distribution along with predicting aesthetic attribute~\cite{zhu2022personalized,zhu2023personalized,shi2024personalized} or personal trait~\cite{li2020personality}. Other GIAA models that utilize score-based comparison~\cite{ke2023vila} or score regression~\cite{ke2021musiq,yi2023towards,zhu2020personalized,li2022transductive} are considered less often.
The fine-tuning for personalization can be improved using several methods. One method uses a meta-learner~\cite{zhu2020personalized,li2022transductive}, either alone or combined with a prior model that predicts personal traits~\cite{zhu2021learning,li2020personality} or aesthetic attributes~\cite{zhu2023personalized,yan2024hybrid}. Importantly, these models take only images as input. 
Other approaches involve models that receive additional personal traits as input~\cite{yang2023multi,zhu2022personalized,shi2024personalized}. 
For example, PIAA-MIR~\cite{zhu2022personalized} and PIAA-ICI~\cite{shi2024personalized} involve learning personal scores by computing the interaction between aesthetic attributes and demographic features for personalization. Specifically, PIAA-ICI goes further by extracting aesthetic attributes from both images and demographic features, constructing separate graphs for each, and computing both internal interactions within each graph and external interactions between the two graphs. In contrast, Multi-Level Transitional Contrast Learning (MCTL)~\cite{yang2023multi} uses contrastive learning to learn trait embeddings from personal aesthetic scores, without explicit demographic features. Unlike PIAA-MIR and PIAA-ICI, which can infer image aesthetics for unseen users without fine-tuning, MCTL cannot generalize to unseen users due to its trait embeddings being tied to specific personal scores.

Despite the success of these approaches, the performance of PIAA models when directly evaluated on unseen users remains unclear, as they are typically evaluated under a meta-learning scheme where the model is fine-tuned on each user. 
While this scheme is appropriate for models that take only images as input~\cite{zhu2020personalized,li2022transductive,yan2024hybrid}, it becomes redundant for models that incorporate personal traits~\cite{zhu2022personalized,shi2024personalized}. These models should ideally infer image aesthetics for unseen users without requiring fine-tuning, thus performing zero-shot inference on unseen users. Moreover, the current evaluation scheme emphasizes performance variation across image sampling while overlooking variation due to user sampling. This work focuses on PIAA models like PIAA-MIR and PIAA-ICI, which are capable of zero-shot inference, and explores performance variation specifically based on demographic factors.

\subsection{Transfer learning}\label{sec:relate:transfer}
Transfer learning leverages prior knowledge from a source task to improve performance on a related target task with limited data~\cite{pan2010survey,zhuang2020comprehensive,tajbakhsh2016convolutional,devlin2018bert,radford2018improving,yun2024scaling}. This approach is essential to existing PIAA research~\cite{zhu2022personalized,zhu2023personalized,li2020personality,shi2024personalized,yun2024scaling}, where personal aesthetic data is typically scarce.
To quantify source–target alignment, task vectors serve as a key metric by capturing the directional shifts in parameter space needed to adapt a model from a source to a target task, thus providing insight into task alignment and suitability for transfer~\cite{ilharco2022editing, achille2019task2vec}; Yun and Choo~\cite{yun2024scaling} utilize task vectors (\ie model parameters of GIAA models) to facilitate the metric comparisons between GIAA datasets. While their approach successfully analyzes multiple GIAA datasets, demographic differences of the individuals across these datasets cannot be further examined using these task vectors. 
Complementary distributional metrics—such as Maximum Mean Discrepancy (MMD) \cite{gretton2012kernel}, Kullback–Leibler (KL) divergence, and Earth Mover’s Distance (EMD) \cite{rubner2000earth}—quantify dataset divergence, while performance-based metrics assess transfer effectiveness by measuring target-task accuracy after source pre-training, thereby guiding optimal model adaptation. Notably, although these metrics have not been used to study transfer learning in existing IAA works, EMD is widely adopted as a training loss in standard IAA protocols~\cite{talebi2018nima,yang2022personalized}.
\section{On the Geometry of Image Aesthetics}\label{sec:geometry}
\noindent{\textbf{Notation}}
Let $s$ represent the aesthetic score. The function $\hat{P}(s)$ denotes the output score distribution generated by an IAA model, which takes images as inputs in a $d$-dimensional space, where $d$ is the number of score intervals. The symbol $\delta_{i}(s)$ represents the groundtruth score distribution for PIAA, expressed as a one-hot vector for an individual score, with $i$ indicating the user.

We demonstrate that for the PIAA models without incorporating personal traits~\cite{zhu2020personalized,li2022transductive,yan2024hybrid}, their performance on PIAA serves as the upper bound for their performance on GIAA, as shown in Theorem~\ref{thm:giaa_piaa_loss}. 
\begin{restatable}{theorem}{giaapiaaloss}
\label{thm:giaa_piaa_loss}
With the notation $\hat{P}(s)$ and $\delta_{i}(s)$ as defined above, and where $n$ is the total number of users, the GIAA and PIAA loss functions are given by
\begin{align}
    \mathcal{L}_{GIAA} &= \left\lVert \hat{P}(s) - \frac{1}{n} \sum_{i=1}^{n} \delta_{i}(s) \right\lVert,\\
    \mathcal{L}_{PIAA} &= \frac{1}{n} \sum_{i=1}^{n} \left\lVert \hat{P}(s) - \delta_{i}(s) \right\lVert.
\end{align}
respectively. Then, we have
\begin{equation}
\mathcal{L}_{GIAA} \leq \mathcal{L}_{PIAA}.
\end{equation}
Note that this result holds not only when $\hat{P}(s)$ and $\delta_{i}(s)$ represent score distributions but also when they are scalar scores.
\end{restatable}
%
% \begin{proof}
% Given $\hat{P}(s)$ is the predicted score distribution by an IAA model, $\delta_{i}(s)$ is the score distribution for user $i$, and $s$ is the score, the GIAA loss function $\mathcal{L}_{GIAA}$ is
% \begin{equation}
% \begin{aligned}
%     \mathcal{L}_{GIAA} &= | \hat{P}(s) - \frac{1}{n} \sum_{i=1}^{n} \delta_{i}(s) | 
%     \\ &= \frac{1}{n} | \sum_{i=1}^{n} ( \hat{P}(s) - \delta_{i}(s) ) | 
%     \\ &\leq \frac{1}{n} \sum_{i=1}^{n} | \hat{P}(s) - \delta_{i}(s) | = \mathcal{L}_{PIAA}
% \end{aligned}
% \end{equation}
% where the inequality holds by the triangular inequality and $\mathcal{L}_{PIAA}$ is the PIAA loss function. This inequality suggests that IAA models perform better on GIAA tasks than on PIAA tasks when the model is unconditioned to the user, even when trained on PIAA data. Note that the same proof applies when predicting scores instead of score distributions. By replacing $\hat{P}(s)$ with the predicted score and $\delta_{i}(s)$ with the score for user $i$, the sketch of proof remains unchanged.
% \end{proof}
%
See the proof in Suppl. Section 1. This theory suggests that IAA models perform better on GIAA tasks than on PIAA tasks when unconditioned on the user. 
%Moreover, i
It follows immediately that PIAA models can generalize well to GIAA data without GIAA pre-training. 

Next, we further show that the same statement holds even when the model is conditioned to the demographic traits (hereafter referred to as traits), \eg, PIAA-MIR~\cite{zhu2022personalized} and PIAA-ICI~\cite{shi2024personalized}. Under this setup, the PIAA models map a pair of images and traits to scores. 
% To make this setup compatible with GIAA, we extend the definition of GIAA such that it maps pairs of mean traits and images to mean scores across users. 
To make this setup compatible with GIAA, we extend the definition of GIAA such that it maps pairs of the averaged traits distribution and images to the average score distribution across users. 
This extension is reasonable because, without accounting for traits, a GIAA model is likely to overfit the training data by simply memorizing preferences linked to the averaged traits distribution. 
This definition provides a clear way to model group preferences.
With this extension, GIAA maps averaged traits distribution to average score distribution, while PIAA maps individual traits to individual scores for given images. 
Moreover, both the input space (traits) and output space (scores) of IAA form distinct convex hulls based on all personal data, as illustrated in Figure~\ref{fig:convex_hull}. %; where they are introduced as trait convex hull and score convex hull, respectively.
% GIAA projects the center of the convex hull in the input space to the center of the convex hull in the output space, 
GIAA projects the average trait distribution located at the inner regime of the trait convex hull to the average score distribution located at the inner regime of the score convex hull,
whereas PIAA projects each vertex from the input convex hull to the corresponding point in the output convex hull. 
This shifts the transfer learning between GIAA and PIAA into a domain generalization problem, revealing the explicit geometry of IAA. Under this framework, transfer learning from GIAA to PIAA can be viewed as extrapolation in the trait domain, while the reverse is interpolation, which is generally more effective in machine learning. Thus, we conclude that PIAA models can generalize well to GIAA data without GIAA pre-training, even when the model is conditioned on users. We provide experimental results to verify this in Section~\ref{sec:results}.

\section{Experimental Results}\label{sec:results}
\subsection{Evaluation Scheme of PIAA}~\label{ssec:results:eval_scheme}
Prior PIAA methods \cite{yang2022personalized,zhu2022personalized,shi2024personalized,zhu2020personalized,li2022transductive,yang2023multi} risk data leakage if the same datasets are used for both pre-training and PIAA fine-tuning. This risk arises because GIAA splits data by images, whereas PIAA divides training and test sets by users, potentially resulting in the same images being present in both the GIAA training phase and the PIAA testing phase. 
To avoid this, we unify the evaluation: all models share a conventional GIAA image split into train/validation/test sets; PIAA uses the full training set (\ie, no few-shot) and is trained collectively across users. For zero-shot generalization to unseen users, we keep the image split but segregate users by demographic (\eg, train on females, test on males). See Suppl. Section 2 for details.

\subsection{Datasets}~\label{ssec:results:dataset}
Despite the abundance of image aesthetics resources in GIAA~\cite{yang2022personalized, kong2016photo, murray2012ava, kang2020eva, he2022rethinking, huang2022aesthetic, Ren_2017_ICCV, yi2023towards}, only a few datasets provide personal aesthetic scores for PIAA, such as FLICK-AES~\cite{Ren_2017_ICCV}, PARA~\cite{yang2022personalized}, and LAPIS~\cite{maerten2025lapis}. In this work, we demonstrate our method on the PARA and LAPIS datasets, which include photos and artworks, respectively. A sample from these datasets is shown in Suppl. Section 3. Notably, FLICK-AES is excluded from this study as it lacks personal trait data.

\noindent \textbf{PARA} dataset~\cite{yang2022personalized} includes 31,220 photos and 438 users—each (image, user) pair includes an aesthetic score and aesthetic attributes. The dataset consists of the demographics of users such as age, gender, education, Big-5 personality traits~\footnote{Openness to experience (O), conscientiousness (C), extraversion (E), agreeableness (A), and neuroticism (N)}, and experience in art and photography. 
We adopt a unified GIAA/PIAA evaluation (Section \ref{ssec:results:eval_scheme}) with 25,398 training, 2,822 validation, and 3,000 test images, and segment users by gender (male/female), age (18–21, 22–25, 26–29, 30–34, 35–40), education (junior high, senior high, technical secondary, junior college, university), and photo/art experience (beginner, competent, proficient, expert).

\noindent \textbf{LAPIS} dataset \cite{maerten2025lapis} contains 11 723 artworks rated by 578 users—each (image, user) pair includes an aesthetic score, art style, demographics (age, gender, education, nationality), and 11 Vienna Art Interest and Art Knowledge (VAIAK) values (VAIAK1-7 and 2VAIAK1-4)~\cite{specker2020vienna}. Images are split into 7,074 train, 2,358 validation, and 2,358 test samples; users are segmented by gender (female/male), age (18–27, 28–38, 39–49, 50–60, 61–71), education (primary, secondary, Bachelor’s, Master’s, Doctorate), nationality (44 countries detailed in Suppl. Section 4), and 2 VAIAK levels (low $\leq 3$ and high $> 3$)

\subsection{Trait Encoding and Models}
\noindent \textbf{Trait Encoding}. 
Unlike existing PIAA work~\cite{zhu2022personalized,shi2024personalized} that uses numeric traits directly, we apply onehot encoding to all traits, including numeric ones like the Big-5 in the PARA dataset and the VAIAKs in the LAPIS dataset. This allows our models to access the full distribution of each trait when computing GIAA, rather than relying on mean values.
For PARA, we encode gender (2), age (5), education (5), photography experience (4), art experience (4), and Big-5 personality (50), yielding 70 dimensions. For LAPIS, we encode gender (4), color-blindness (2), age (5), education (5), nationality (44), and VAIAK art-interest/knowledge scores (77), totaling 137 dimensions. By comparison, conventional encodings produce just 25 dimensions for PARA and 71 for LAPIS (see Suppl. Section 6).

\noindent \textbf{Models.} 
As outlined in Section \ref{sec:relate}, we reimplemented PIAA-MIR and PIAA-ICI as our PIAA baselines—reproducing their results (Suppl. Section 5)—and adopt NIMA \cite{talebi2018nima} as our GIAA baseline. We choose NIMA over newer GIAA models \cite{he2022rethinking,yi2023towards} because our PIAA models build on NIMA and the newer models’ use of extra attributes (theme, style) would preclude a fair comparison.
We adapt NIMA, PIAA-MIR, and PIAA-ICI to onehot encoding, enabling inference for both GIAA and PIAA, which we refer to as NIMA-trait, PIAA-MIR (Onehot enc.), and PIAA-ICI (Onehot enc.), respectively. For NIMA-trait, these traits are integrated into NIMA's predictions via an additional two-layer multilayer perceptron (MLP) as detailed in Suppl. Section 6. 
These models are built on pretrained backbones, including ResNet-50~\cite{he2015deep}, Swin-Tiny~\cite{liu2021swin}, and ViT-Small~\cite{dosovitskiy2020image}. For the GIAA inference of NIMA-trait, PIAA-MIR (Onehot enc.), and PIAA-ICI (Onehot enc.), \textbf{we adjust the inference method so that models receive the average trait distribution across all training users during evaluation.} This ensures a fair comparison to the GIAA scenario, where images are the only input.
We follow the standard IAA training protocol~\cite{talebi2018nima,yang2022personalized}, training these models with EMD loss and evaluating them using SROCC and PLCC. 

\subsection{Model Evaluation on overlapped users}
\begin{figure}[h]
    \centering
    \includegraphics[width=1.\linewidth]{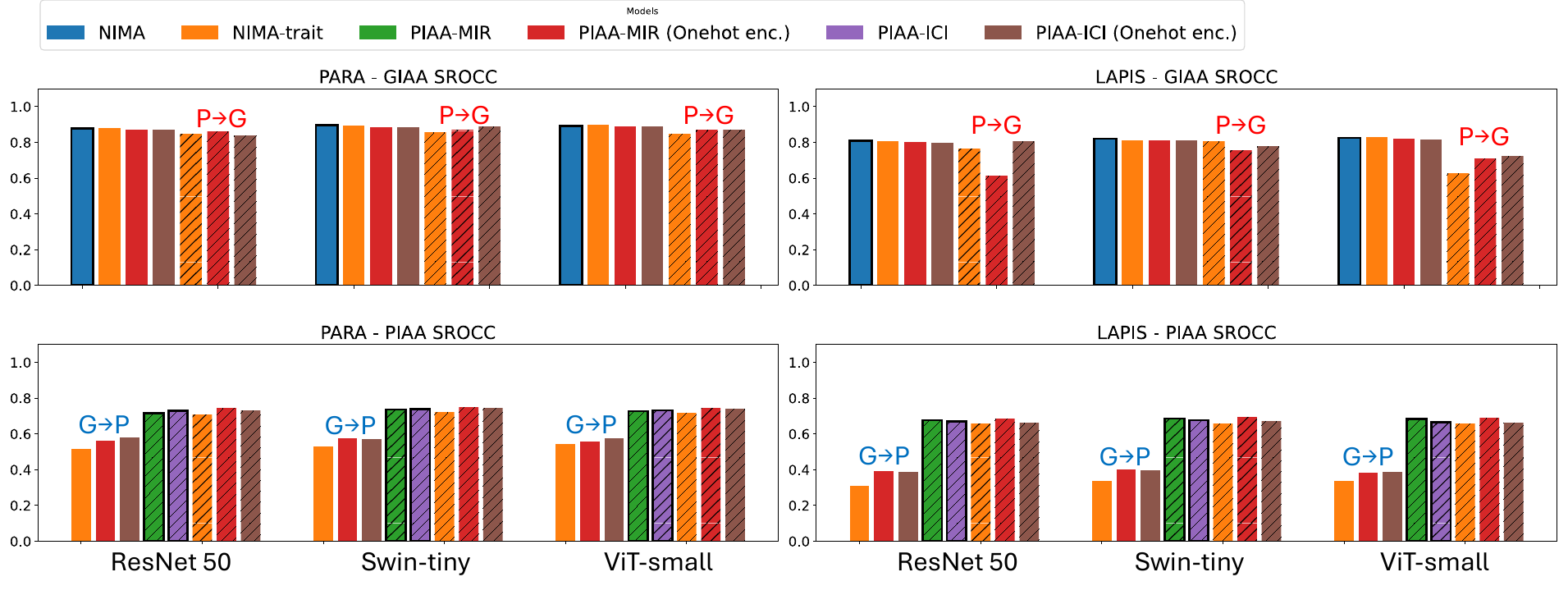}
    \caption{\textbf{GIAA/PIAA SROCC of IAA models using the PARA and LAPIS datasets.} Hatched bars indicate models trained on PIAA data, solid bars those trained on GIAA data, and bars with a black outline denote the respective GIAA/PIAA baselines. Zero-shot PIAA performance (trained on GIAA, tested on PIAA) is shown as \textcolor{blue}{$G\!\rightarrow\!P$}, while zero-shot GIAA performance (trained on PIAA, tested on GIAA) is shown as \textcolor{red}{$P\!\rightarrow\!G$}.
    }    
    \label{fig:table1}
    % \vspace{-0.2cm}
\end{figure}

The SROCC performance of NIMA, PIAA-ICI, PIAA-MIR, and their corresponding onehot-encoded models are shown in Figure~\ref{fig:table1}. The numeric values for both SROCC and PLCC are demonstrated in Suppl. Section 7.
When trained on GIAA, onehot-encoded models achieves performance comparable to the NIMA baseline.
% \footnote{Note that the reported SROCC for NIMA on the PARA dataset is 0.8790~\cite{yang2022personalized}, and our implementation achieves a comparable score.}.
On the other hand, \textbf{when trained on PIAA, these models even outperform the PIAA baselines such as PIAA-MIR and PIAA-ICI, which require GIAA pre-training.} These results demonstrate that our simple yet effective approach performs well in both the GIAA and PIAA settings.
Moreover, the generalization between GIAA and PIAA is evident, where the zero-shot PIAA performance (\textcolor{blue}{$G\rightarrow P$}) is significantly worse than the PIAA baselines while the zero-shot GIAA performance (\textcolor{red}{$P\rightarrow G$}) is generally comparable to the GIAA baseline, with only a few exceptions observed on the LAPIS dataset. This observation aligns with our analysis in Section~\ref{sec:geometry}.

\begin{figure}[h]
    \centering
    \includegraphics[width=1.\linewidth]{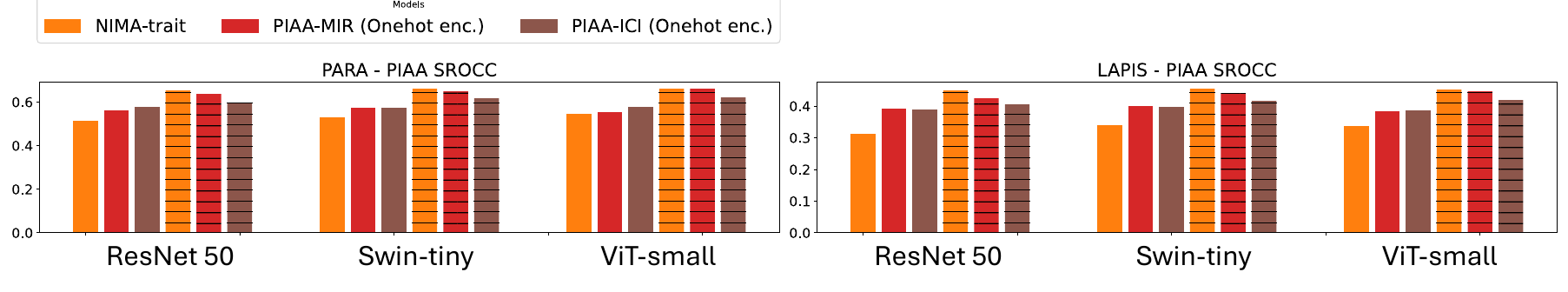}
    \caption{\textbf{Zero-shot PIAA SROCC of IAA models using the PARA and LAPIS datasets.} Hatched bars indicate models trained on sGIAA data and solid bars those trained on GIAA data. The performance improves markedly when using sGIAA as augmentation compared to training on GIAA alone.}
    \label{fig:table2}
    % \vspace{-0.6cm}
\end{figure}

Building on the generalization discussed above, the number of users involved in GIAA likely affects generalization, as fewer users shifts the training domain toward individual preferences, while more users provides greater confidence in the annotated scores. Although existing works have explored GIAA, the effect of group size has rarely been addressed. To investigate this, we propose sub-sampling the GIAA dataset, referred to as sGIAA, with user groups ranging from 2 to the maximum number of users per image, as a form of data augmentation. The model performances in SROCC for this approach are shown in Figure~\ref{fig:table2}, while the numeric values for both SROCC and PLCC are demonstrated in Suppl. Section 7.
These results demonstrate that \textbf{sub-sampling significantly improves zero-shot PIAA performance} by up to 44\% while maintaining GIAA performance. This reveals how sub-sampling helps strike a balance between individual and group preferences, further emphasizing the importance of the number of users in the GIAA dataset.

\subsection{Model Evaluation on Disjoint Users across Demography}
\begin{figure*}
  \captionsetup{font=small} % apply only inside this figure
  \begin{minipage}[b]{1.0\linewidth}
    \begin{subfigure}{0.2\linewidth}
      \includegraphics[width=0.85\linewidth]{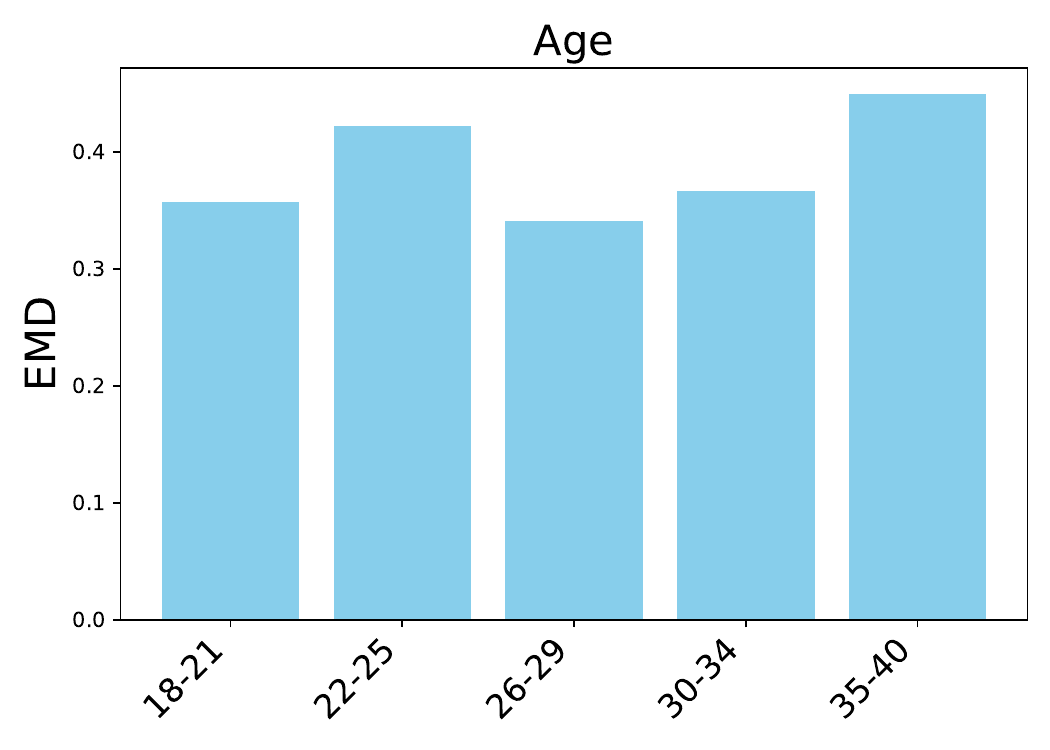}
      \centering
      \caption{PR-Age}\label{fig:PR-age}
    \end{subfigure}\hfill
    \begin{subfigure}{0.2\linewidth} 
      \includegraphics[width=0.85\linewidth]{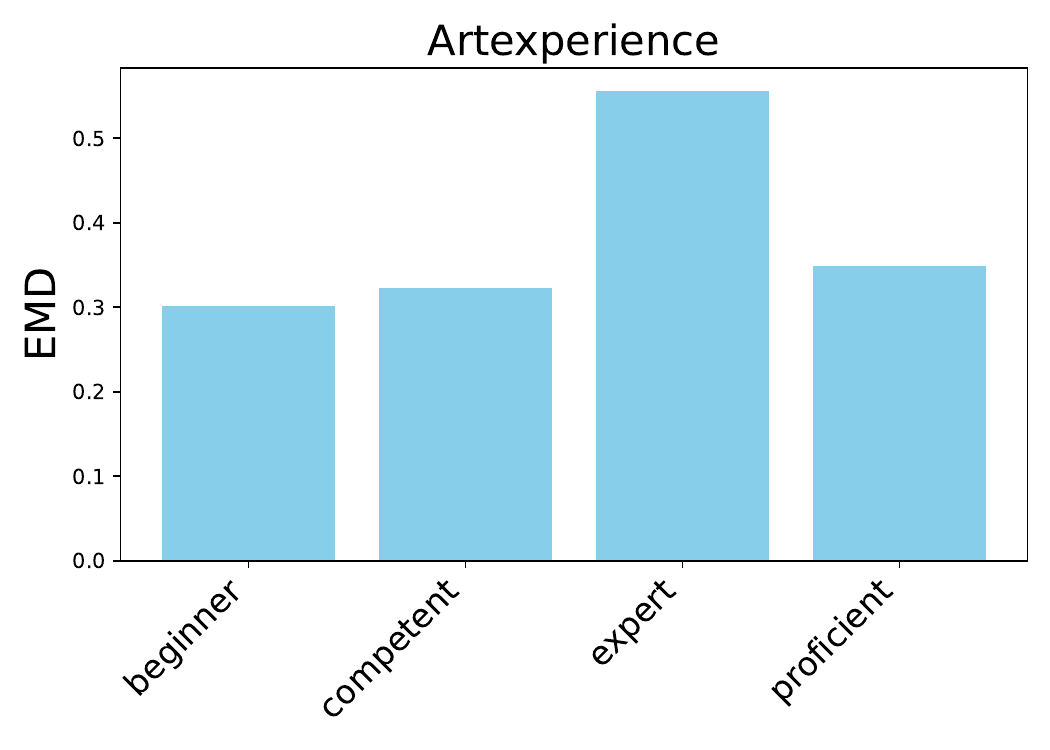}
      \centering
      \caption{PR-ArtExp.}\label{fig:PR-artexperience}
    \end{subfigure}\hfill
    \begin{subfigure}{0.2\linewidth} 
      \includegraphics[width=0.85\linewidth]{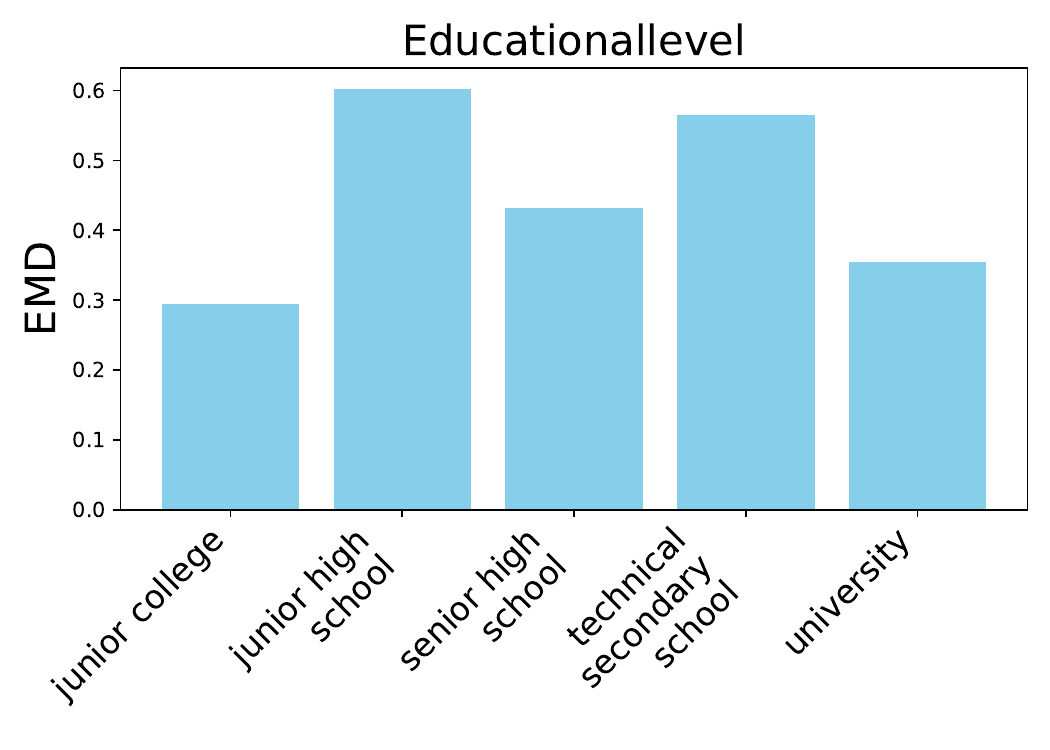}
      \centering
      \caption{PR-Education}\label{fig:PR-educationallevel}
    \end{subfigure}\hfill
    \begin{subfigure}{0.2\linewidth} 
      \includegraphics[width=0.85\linewidth]{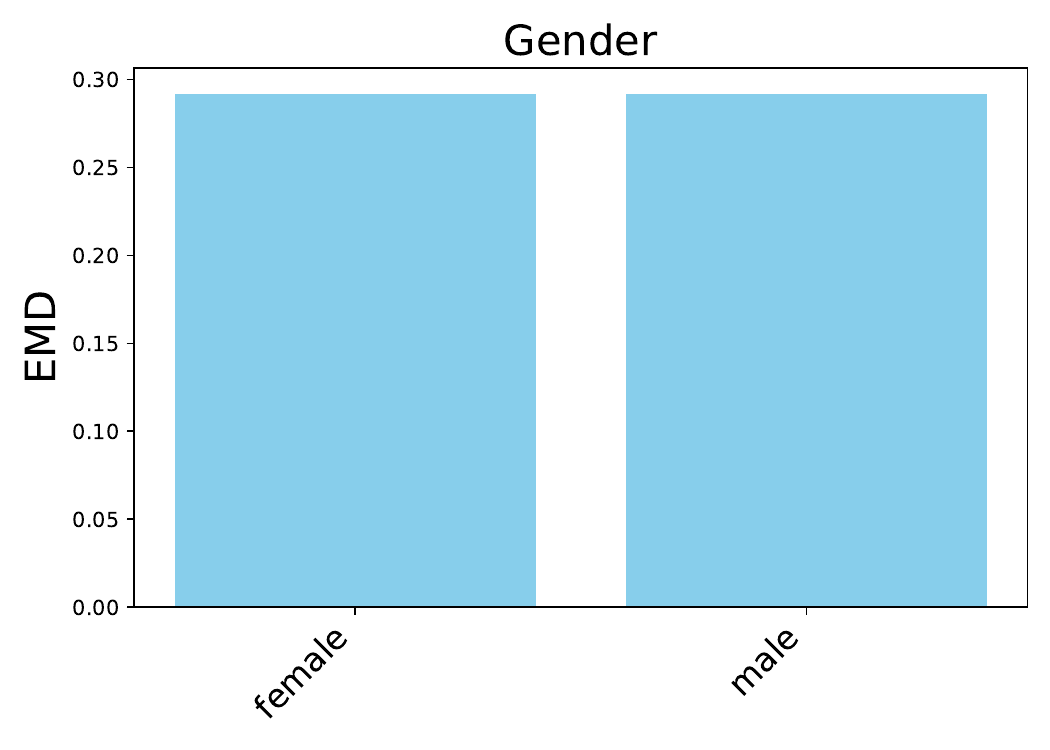}
      \centering
      \caption{PR-Gender}\label{fig:PR-gender}
    \end{subfigure}\hfill
    \begin{subfigure}{0.2\linewidth} 
      \includegraphics[width=0.85\linewidth]{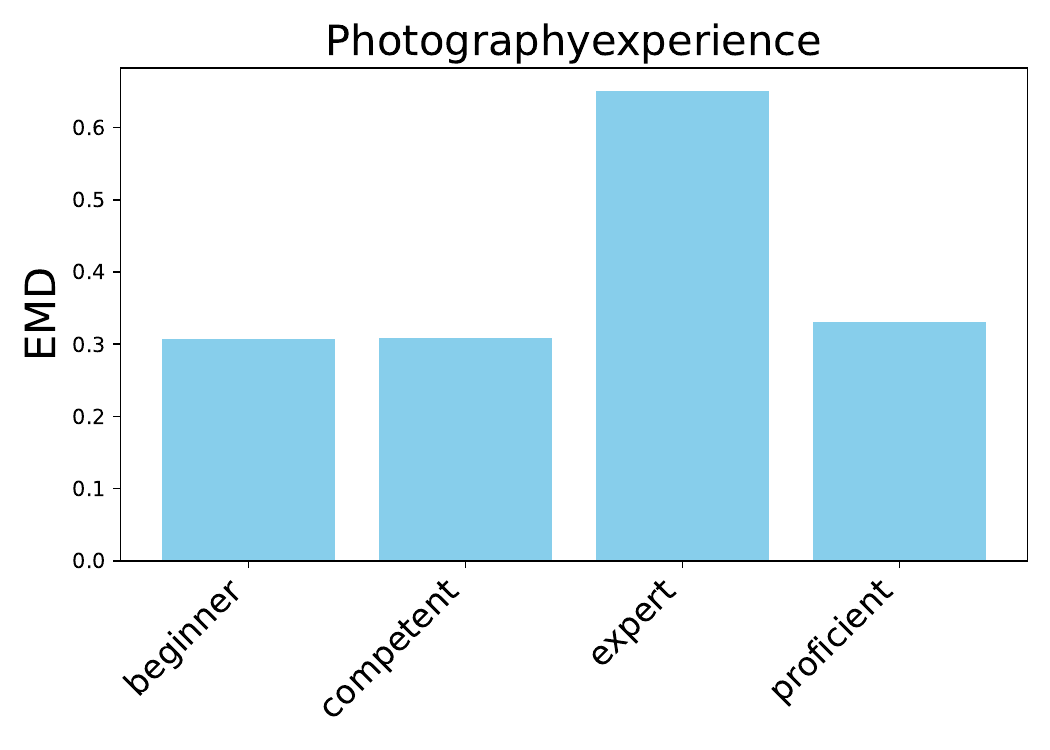}
      \centering
      \caption{PR-PhotoExp.}\label{fig:PR-photographyexperience}
    \end{subfigure}\hfill
    \begin{subfigure}{0.2\linewidth} 
      \includegraphics[width=0.85\linewidth]{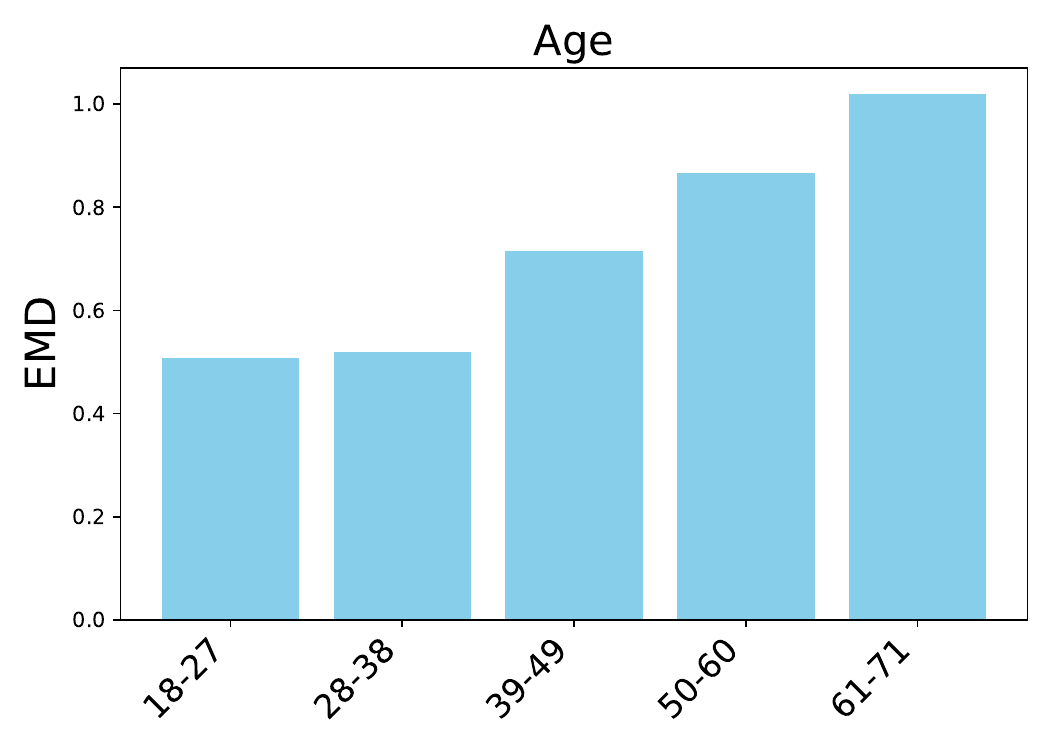}
      \centering
      \caption{LAPIS-Age}\label{fig:lapis-age}
    \end{subfigure}\hfill
    \begin{subfigure}{0.2\linewidth}
      \includegraphics[width=0.85\linewidth]{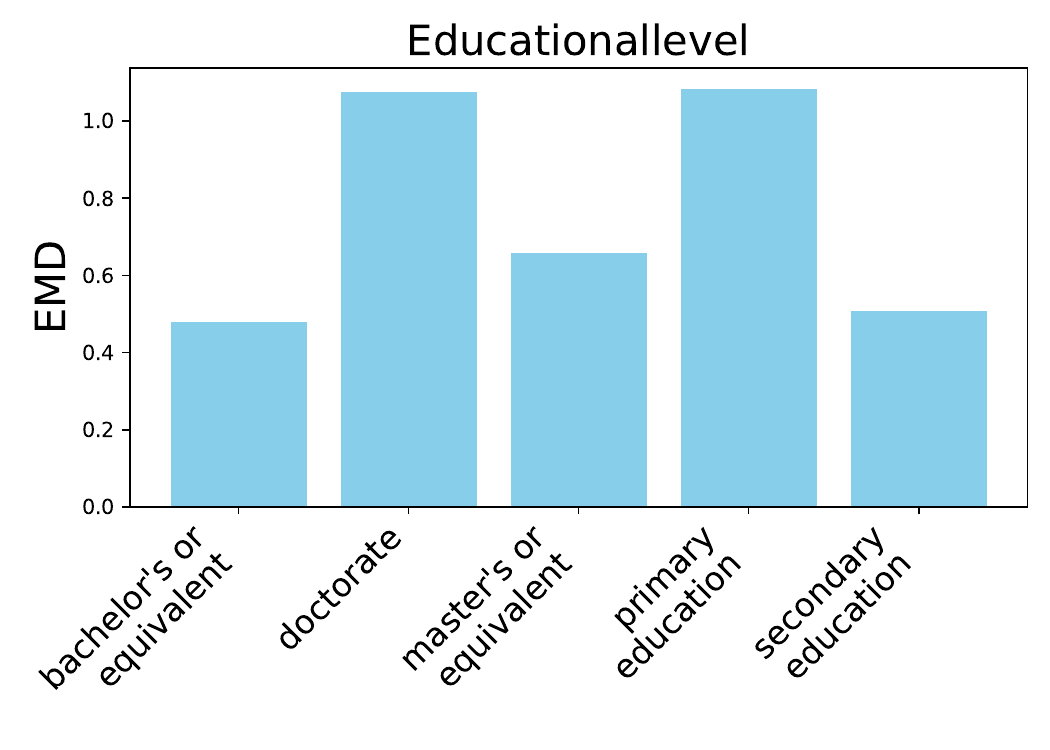}
      \centering
      \caption{LAPIS-Education}\label{fig:lapis-educationallevel}
    \end{subfigure}\hfill
    \begin{subfigure}{0.2\linewidth}
      \includegraphics[width=0.85\linewidth]{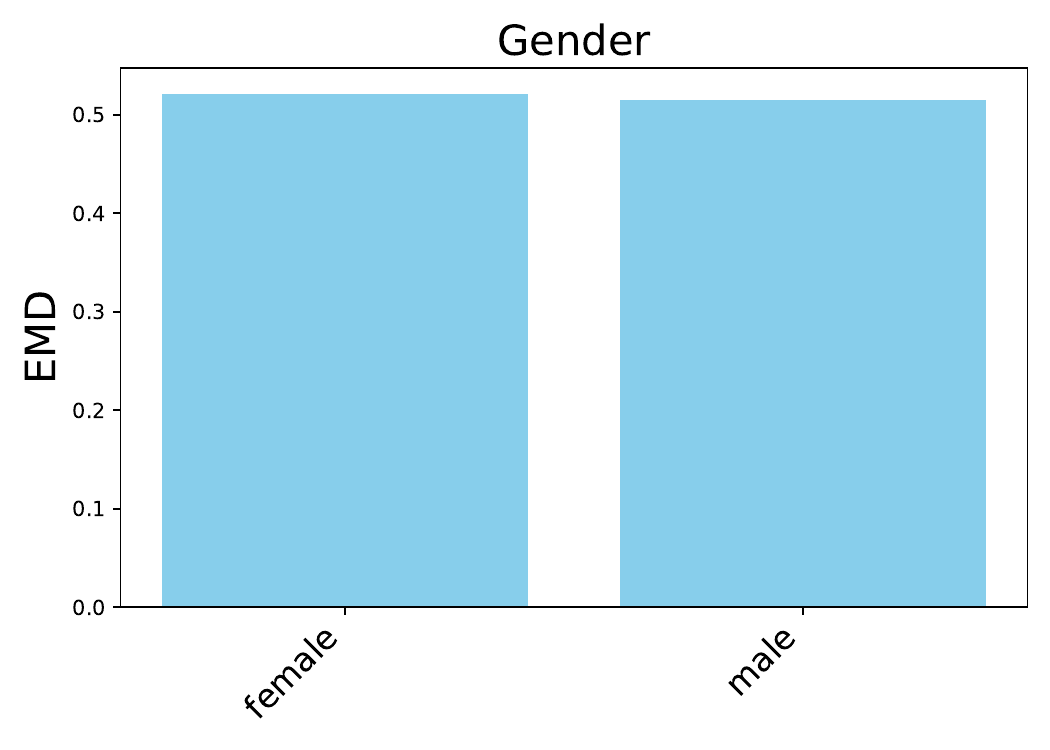}
      \centering
      \caption{LAPIS-Gender}\label{fig:lapis-gender}
    \end{subfigure}\hfill
    \begin{subfigure}{0.2\linewidth}
      \includegraphics[width=0.85\linewidth]{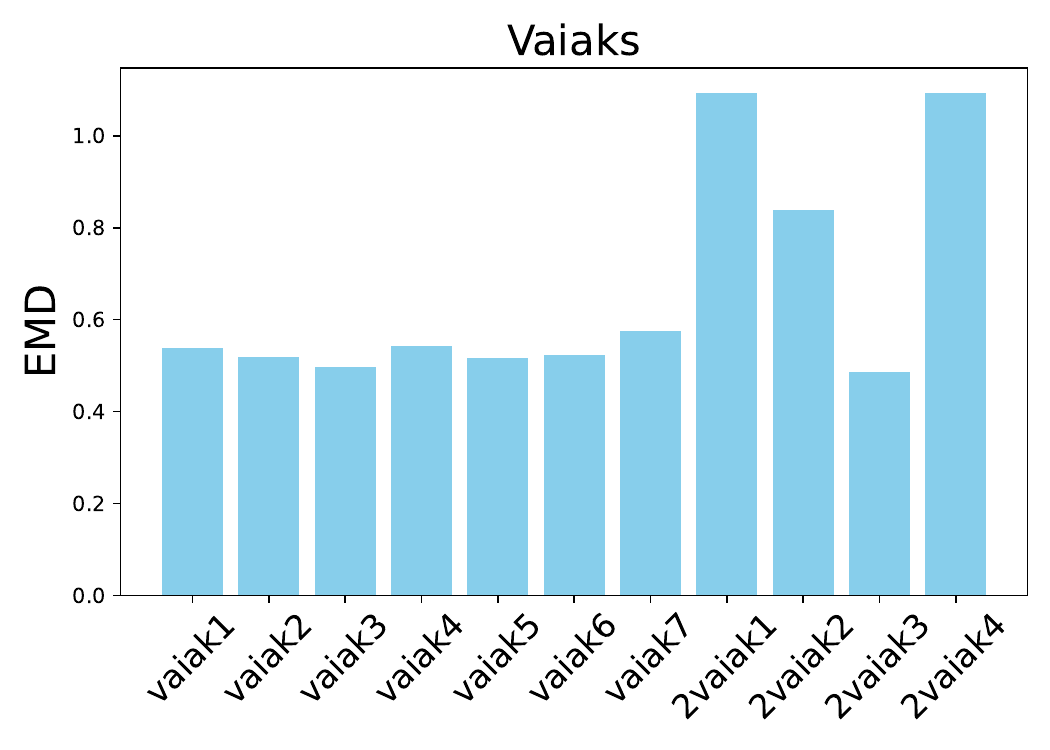}
      \centering
      \caption{LAPIS-VAIAKs}\label{lapis-vaiak}
    \end{subfigure}
    \caption{EMD between disjoint users split by demographic groups on PARA (a–e) and LAPIS (f–i) datasets. “PR” denotes the PARA dataset.}
    \vspace{-0.4cm}
    \label{fig:emd}
  \end{minipage}
\end{figure*}

\noindent\textbf{Analysis of demographic differences.} To further assess the aesthetic differences and model generalization across the demographic split, we select users with a specific trait (e.g., users aged 18-21) as the test users, while all other users serve as the training users as described in section~\ref{ssec:results:eval_scheme}. We hereafter refer to this setup as the disjoint user split.
We then compute the Earth Mover's Distance (EMD) between the aesthetic score distributions of the train and test groups for various demographic splits, as shown in Figure~\ref{fig:emd}. A higher EMD indicates a greater distinction in the aesthetic preferences of the test users compared to the training users. 
The EMD values split by gender are the lowest, while splits based on art experience, photography experience, and educational level show higher EMD values, reaching up to around 0.8. Specifically, experts in both photography and art, as well as users with only high school education, demonstrate the greatest aesthetic distinction.
For the LAPIS dataset, splits based on age, educational level, 2VAIAK1, and 2VAIAK4 yield even higher EMD values, reaching up to approximately 1.2. Overall, EMD values in the LAPIS dataset are higher than those in the PARA dataset, suggesting that aesthetic preferences for artworks are more subjective than for photographs—consistent with previous findings~\cite{vessel2018stronger}.
% In particular, older users, individuals with either a doctorate or primary education, and those with higher art experience exhibit the most distinct aesthetic preferences for artworks.
We also compute the Gini index to quantify the relative importance of each demographic factor. The results are consistent with the EMD analysis, indicating that education level and photography/art experience are the strongest distinguishing factors in the score distributions (Suppl. Section 8).

\begin{figure}[t]
    \centering
    \begin{subfigure}[b]{0.45\linewidth}
        % \centering
        \includegraphics[width=\linewidth]{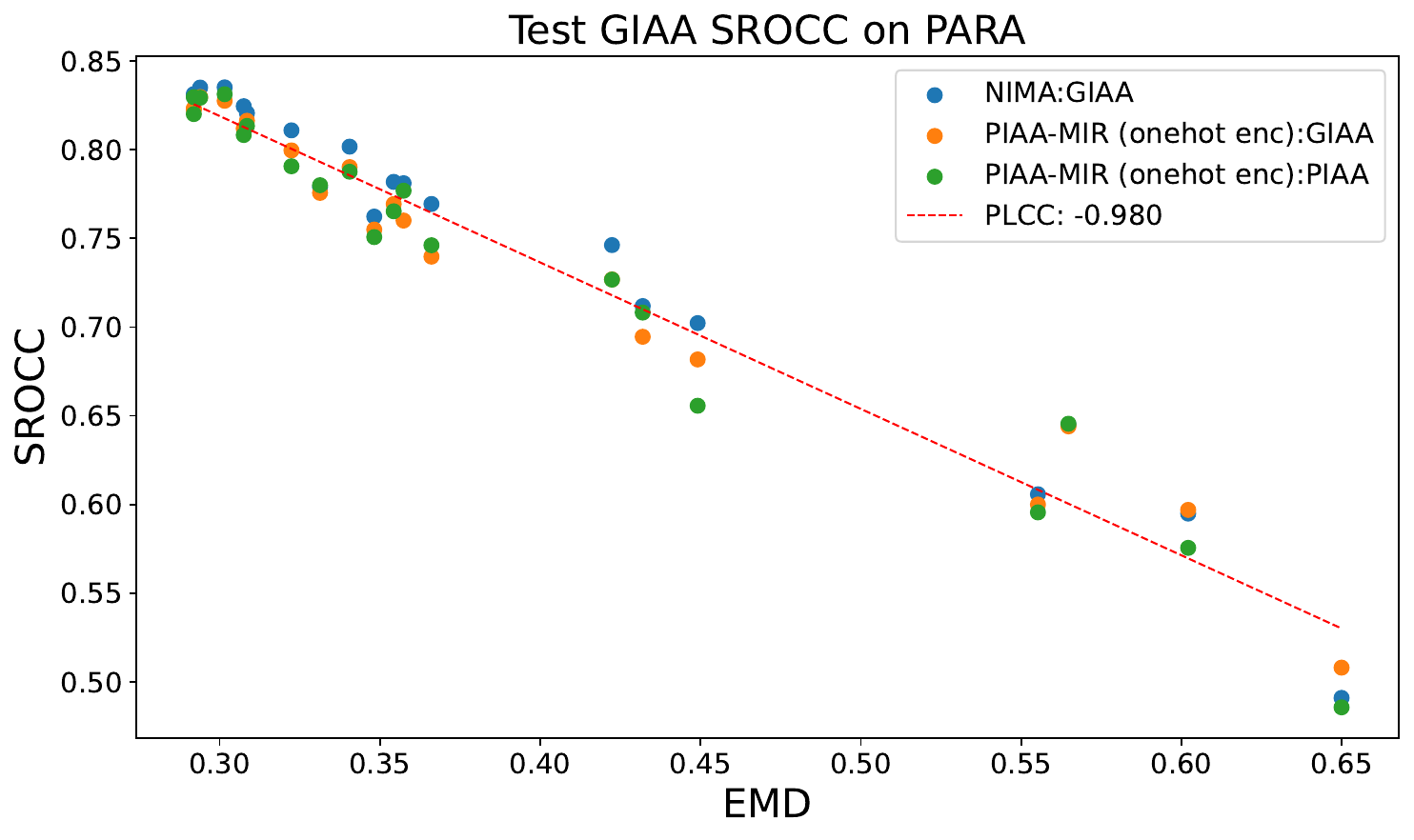}
        \caption{PARA dataset}
        \label{fig:giaa:para}
    \end{subfigure}
    % \hfill
    \begin{subfigure}[b]{0.45\linewidth}
        % \centering
        \includegraphics[width=\linewidth]{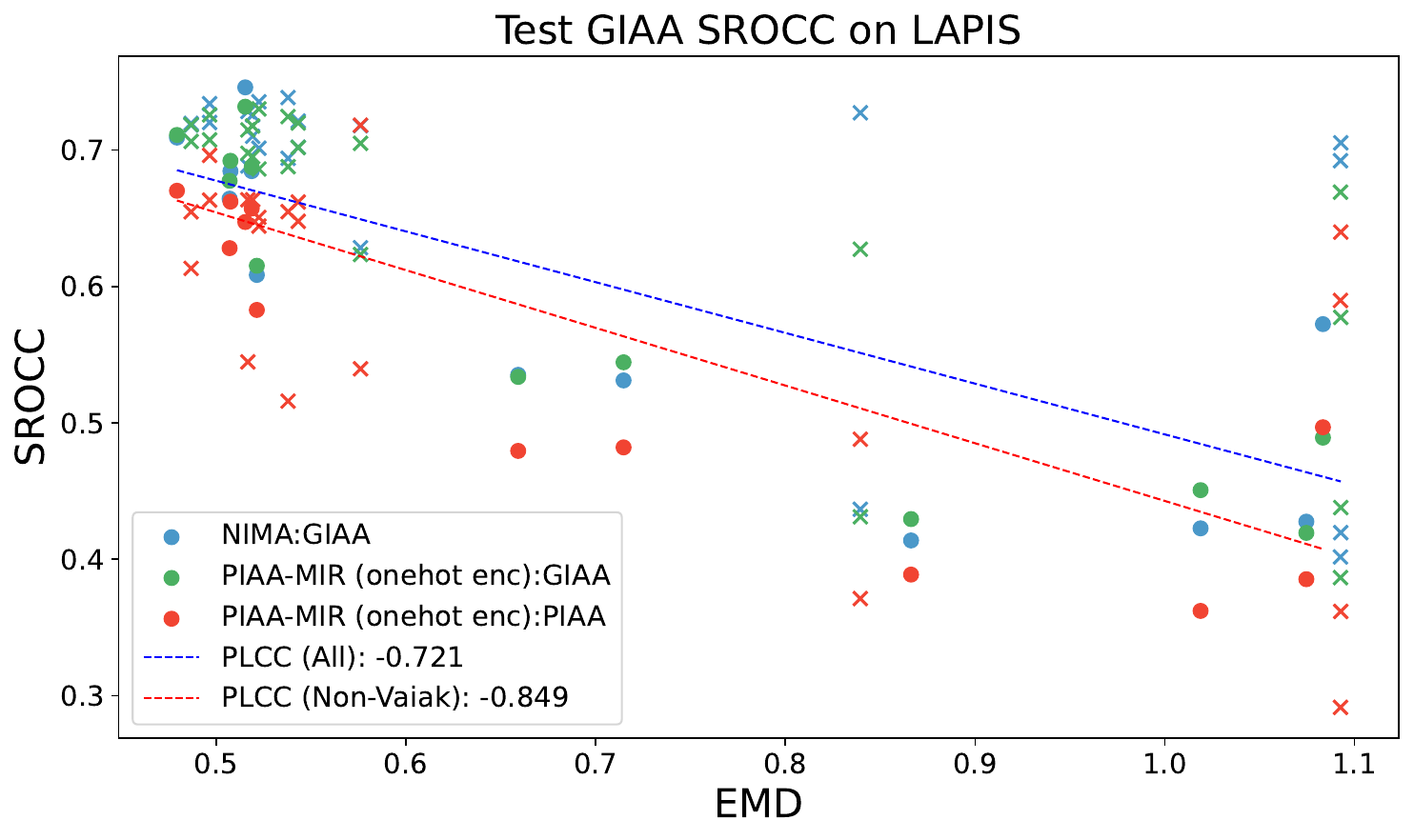}
        \caption{LAPIS dataset. }
        \label{fig:giaa:lapis}
    \end{subfigure}
    \caption{Correlation between GIAA SROCC and EMD across various demographic splits. For LAPIS dataset, the symbol 'x' denotes data split by VAIAKs, with the same color as the 'o' symbols representing the same model.
    }
    % \vspace{-0.4cm}
    \label{fig:giaa}
\end{figure}

\begin{figure}[t]
    \centering
    \begin{subfigure}[b]{0.45\linewidth}
        \centering
        \includegraphics[width=\linewidth]{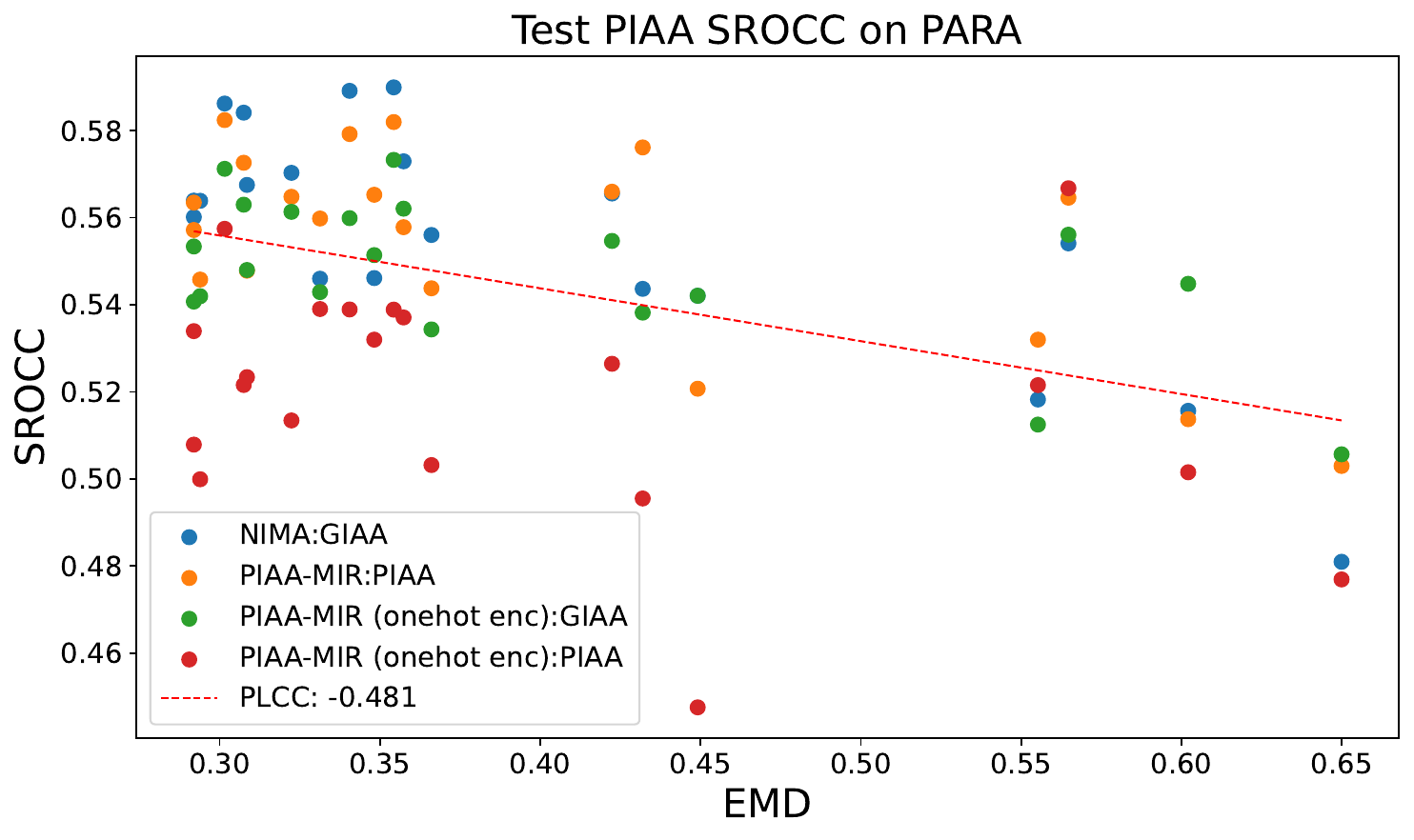}
        \caption{PARA dataset}
        \label{fig:piaa:para}
    \end{subfigure}
    % \hfill
    \begin{subfigure}[b]{0.45\linewidth}
        \centering
        \includegraphics[width=\linewidth]{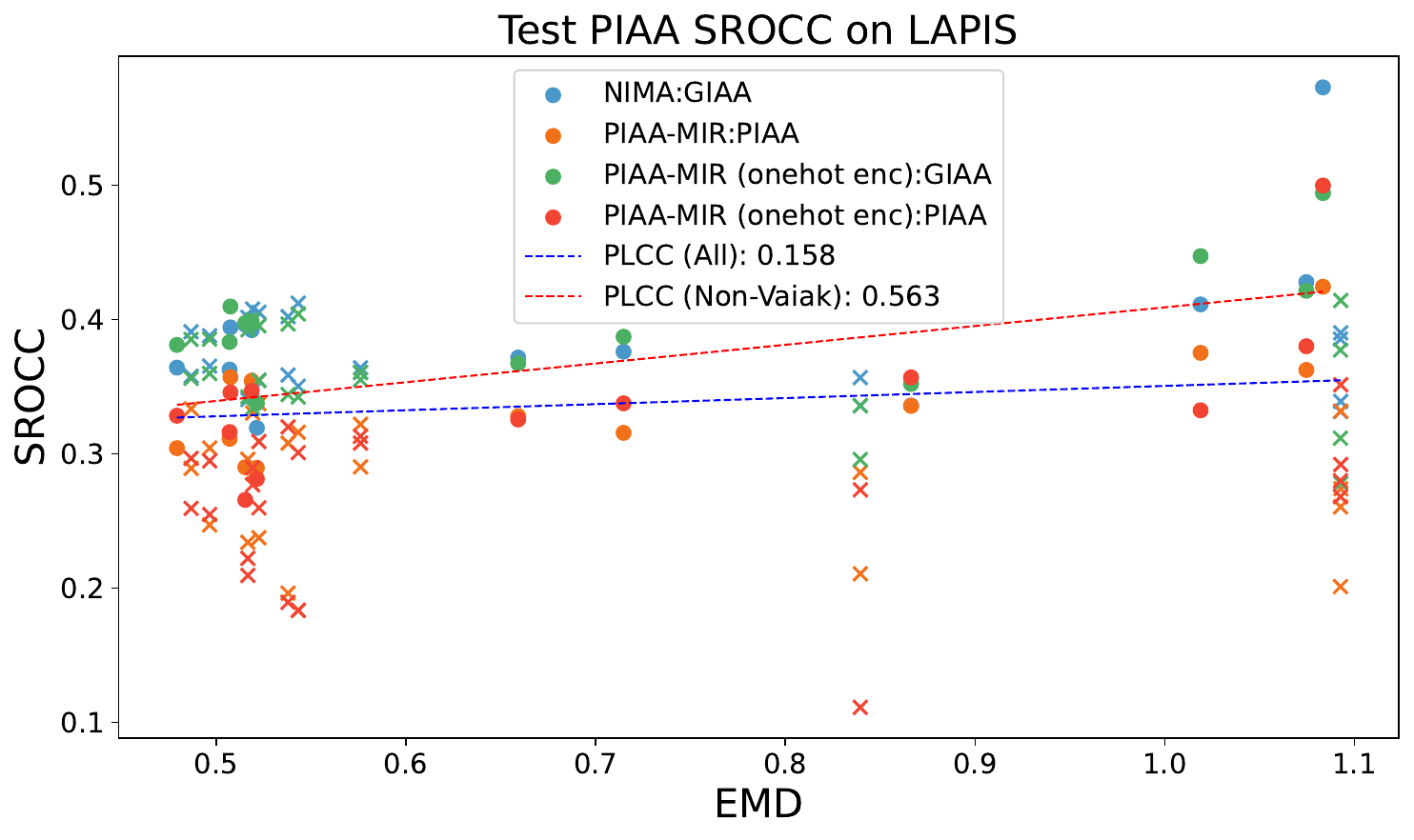}
        \caption{LAPIS dataset}
        \label{fig:piaa:lapis}
    \end{subfigure}
    \caption{Correlation between PIAA SROCC and EMD across various demographic splits. The format follows that of Figure~\ref{fig:giaa}.}
    % \vspace{-0.4cm}
    \label{fig:piaa}
\end{figure}

\noindent\textbf{Generalization to new users.} Using the disjoint user split, we evaluate NIMA, PIAA-MIR, and PIAA-MIR (Onehot-enc.) across various demographic splits, focusing on models with a ResNet-50 backbone. Further details are provided in Suppl. Section 9.
In Figure~\ref{fig:giaa}, we plot GIAA SROCC against EMD to show that aesthetic differences driven by demographics are substantial across all models. Performance varies markedly, with SROCC ranging from 0.486 to 0.835 on the PARA dataset and from 0.292 to 0.746 on the LAPIS dataset—equivalent to performance gaps of 41.8\% and 60.9\%, respectively. \textbf{These performance differences across demographic groups indicate systematic variation in aesthetic preferences, showing that even aggregated group scores cannot fully remove individual subjectivity.}
Furthermore, the results reveal a strong negative correlation between aesthetic distinction (EMD) and model generalization (SROCC), with PLCCs of –0.980 for PARA and –0.721 for LAPIS. Removing demographic outliers in LAPIS—where certain VAIAK groups are highly imbalanced—improves the PLCC to –0.849. \textbf{These results demonstrate that EMD is an effective metric for estimating GIAA model generalization to unseen users.}
% This high PLCC highlights intrinsic aesthetic differences even for GIAA, suggesting that existing GIAA methods may overlook demographic differences. It is also worth noting that class imbalance is more pronounced in the LAPIS dataset compared to the PARA dataset. For instance, users with primary education and doctorate degrees make up only 0.9\% and 1.5\% of the total data in LAPIS, respectively. In contrast, the rarest category in PARA—users with photography expertise—accounts for 2.0\% of the data, while all other demographic groups exceed 7\%. This greater imbalance in user demographics in LAPIS contributes to more outliers in the analysis. 
A similar analysis on PIAA is conducted and illustrated in Figure~\ref{fig:piaa}. We observe a smaller variation in SROCC values for the PARA dataset, ranging from 0.448 to 0.590, while the LAPIS dataset exhibits a significantly larger variation, with SROCC values ranging from 0.111 to 0.573. Namely, the model performance can vary by as much as 24.1\% and 80.6\% on the two datasets, respectively. 
The greater variation in performance on the LAPIS dataset further underscores the higher individual subjectivity associated with artworks compared to photographs in both GIAA and PIAA models. In particular, the LAPIS dataset shows stronger performance variation for PIAA than for GIAA, indicating a greater challenge in achieving model generalization for unseen users with diverse demographic profiles. The weaker negative correlation between PIAA SROCC values and EMD, compared to GIAA, may indicate improved generalization to unseen users. 

\section{Conclusion}
%In summary, 
We propose the first model capable of supporting both GIAA and PIAA, matching GIAA baseline performance and even surpassing state-of-the-art PIAA models that require GIAA pre-training. 
For GIAA, while our method requires additional trait data during training, it demonstrates efficacy by using the averaged trait distribution from the training data as a fixed input, enabling the model to rely solely on image inputs during inference. Additionally, our model introduces the first theoretical framework for addressing aesthetic differences between groups and individuals, accounting for diverse demographics and group size.
Our comprehensive experiments investigate the transfer learning between GIAA and PIAA under these factors. The results support our theory that transferring from GIAA to PIAA involves extrapolation, while the reverse—interpolation—is generally more effective for machine learning. Additionally, sub-sampled GIAA (sGIAA) improves zero-shot PIAA performance by 44\%, underscoring the importance of group size variation, especially for PIAA fine-tuning in the current scenario. 
For unseen users from diverse demographics, large performance variations highlight the limited generalizability of IAA models and challenge the long-standing assumption that score averaging suppresses subjectivity. We also present the first quantitative evidence that artworks exhibit greater subjectivity than photographs.
% \appendix
\section*{Acknowledgment}
This research was funded by the European Union (ERC Advanced Grant GRAPPA, 101053925, awarded to Johan Wagemans). We thank Yongzhen Ke for providing the pretrained PIAA-ICI model.
\bibliography{egbib}

\end{document}

\maketitle

\section{The Correlation between GIAA and PIAA Performance}\label{sec:som:giaa_upbound}

\renewcommand{\thetheorem}{3.1}
\begin{theorem}
\label{thm:giaa_piaa_loss}
Let $s$ represent the aesthetic score, $\hat{P}(s)$ denote the predicted score distribution produced by an IAA model that takes only images as input, and $\delta_{i}(s)$ denote the ground-truth score distribution for user $i$, expressed as a one-hot vector for an individual score. Let $n$ be the total number of users.

The GIAA and PIAA loss functions are defined as
\begin{align}
    \mathcal{L}_{GIAA} &= \left\lVert \hat{P}(s) - \frac{1}{n} \sum_{i=1}^{n} \delta_{i}(s) \right\lVert,\\
    \mathcal{L}_{PIAA} &= \frac{1}{n} \sum_{i=1}^{n} \left\lVert \hat{P}(s) - \delta_{i}(s) \right\lVert,
\end{align}
respectively. Then, we have
\begin{equation}
\mathcal{L}_{GIAA} \leq \mathcal{L}_{PIAA}.
\end{equation}
This result holds not only when $\hat{P}(s)$ and $\delta_{i}(s)$ represent score distributions but also when they are scalar scores.
\end{theorem}

\begin{proof}
Given $\hat{P}(s)$ is the predicted score distribution by an IAA model, $\delta_{i}(s)$ is the score distribution for user $i$, and $s$ is the score, the GIAA loss function $\mathcal{L}_{GIAA}$ is
\begin{equation}
\begin{aligned}
    \mathcal{L}_{GIAA} &= \lVert \hat{P}(s) - \frac{1}{n} \sum_{i=1}^{n} \delta_{i}(s) \lVert 
    \\ &= \frac{1}{n} \lVert \sum_{i=1}^{n} ( \hat{P}(s) - \delta_{i}(s) ) \lVert 
    \\ &\leq \frac{1}{n} \sum_{i=1}^{n} \lVert \hat{P}(s) - \delta_{i}(s) \lVert = \mathcal{L}_{PIAA}
\end{aligned}
\end{equation}
where the inequality holds by the triangular inequality and $\mathcal{L}_{PIAA}$ is the PIAA loss function. This inequality suggests that IAA models perform better on GIAA tasks than on PIAA tasks when the model is unconditioned to the user, even when trained on PIAA data. Note that the same proof applies when predicting scores instead of score distributions. By replacing $\hat{P}(s)$ with the predicted score and $\delta_{i}(s)$ with the score for user $i$, the sketch of proof remains unchanged.
\end{proof}

\section{Data Splitting for IAA}\label{sec:som:data_split}
\begin{figure}[ht]
    \centering
    % First subfigure
    \begin{subfigure}[]{\linewidth}
        \centering
        \includegraphics[width=0.6\linewidth]{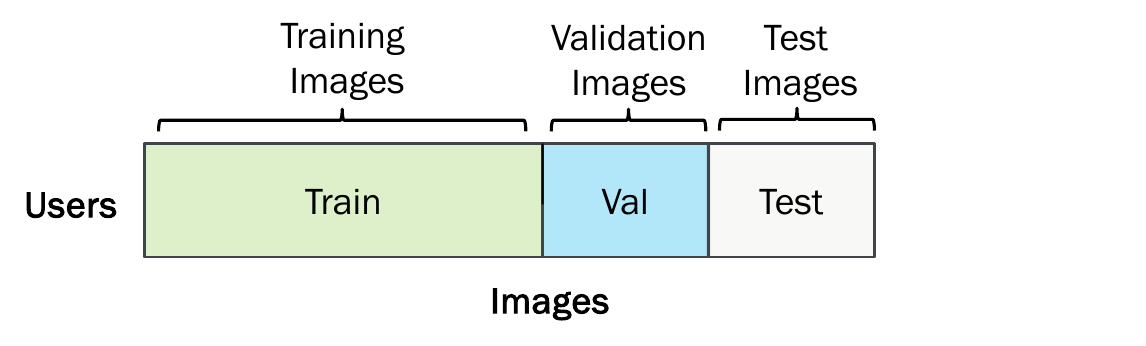}
        \caption{Data Splitting for GIAA. The training, validation, and test data are split according to images.}
        \label{fig:data_split:giaa}
    \end{subfigure}
    % Second subfigure
    \begin{subfigure}[]{\linewidth}
        \centering
        \includegraphics[width=0.6\linewidth]{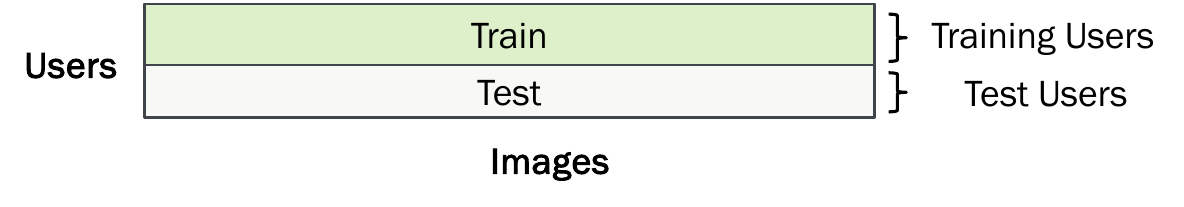}
        \caption{Data Splitting for PIAA. The training and test data are split according to users, followed by few-shot image sampling for training and evaluation.}
        \label{fig:data_split:piaa}
    \end{subfigure}
    % Third subfigure
    \begin{subfigure}[]{\linewidth}
        \centering
        \includegraphics[width=0.6\linewidth]{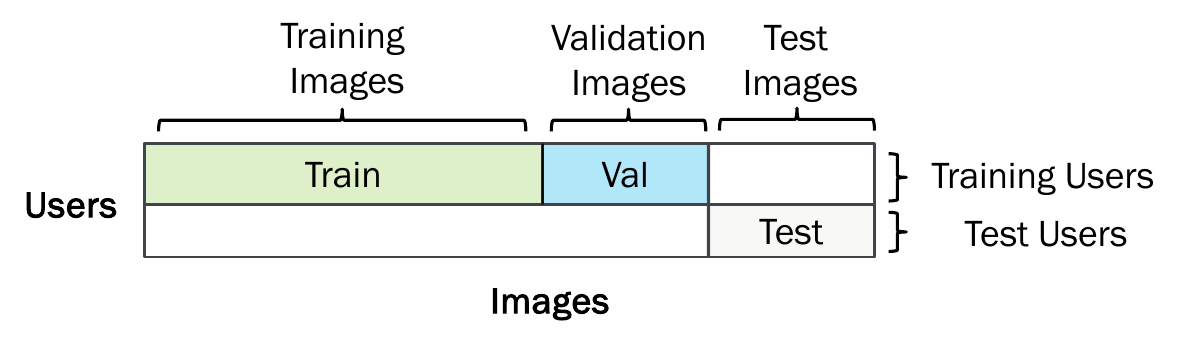}
        \caption{Data Splitting for the zero-shot evaluation scheme on unseen users. The training set comprises (training images, training users), the validation set includes (validation images, training users), and the test set contains (test images, test users).}
        \label{fig:data_split:consistent}
    \end{subfigure}
    \vspace{0.1cm}
    \caption{An overview of the data splitting for (a) GIAA, (b) PIAA, and (c) our proposed consistent splitting for both GIAA and PIAA to prevent data leakage. \textit{Train}, \textit{Val}, and \textit{Test} represent training, validation, and test data, respectively.}
    \label{fig:data_split}
\end{figure}

Existing PIAA studies~\cite{yang2022personalized,zhu2022personalized,shi2024personalized,zhu2020personalized,li2022transductive,yang2023multi} rely on pre-trained GIAA models, which can introduce data leakage when the same datasets are employed for both GIAA pre-training and PIAA fine-tuning. 
The data leakage issue arises because GIAA models typically split data by images, as shown in Figure~\ref{fig:data_split:giaa}, whereas PIAA separates training and test sets by users, as illustrated in Figure~\ref{fig:data_split:piaa}, followed by few-shot sampling. As a result, the same images may appear in both the GIAA training phase and the PIAA testing phase.

In this work, we adopt an alternative data-splitting strategy to evaluate models' zero-shot performance on unseen users, addressing data leakage and establishing a consistent evaluation framework for both GIAA and PIAA. We follow the standard GIAA approach by initially dividing the data into training, validation, and test images. Additionally, we separate training and test users based on demographics; for instance, females are selected as training users, while males are designated as test users. Specifically, the training set comprises (training images, training users), the validation set includes (validation images, training users), and the test set contains (test images, test users), as shown in Figure~\ref{fig:data_split:consistent}.

\section{Example from the PARA and LAPIS Datasets}\label{sec:som:datasets}
\begin{figure}[h]
    \centering
    \begin{subfigure}[b]{0.48\textwidth}
        \centering
        \includegraphics[width=\linewidth]{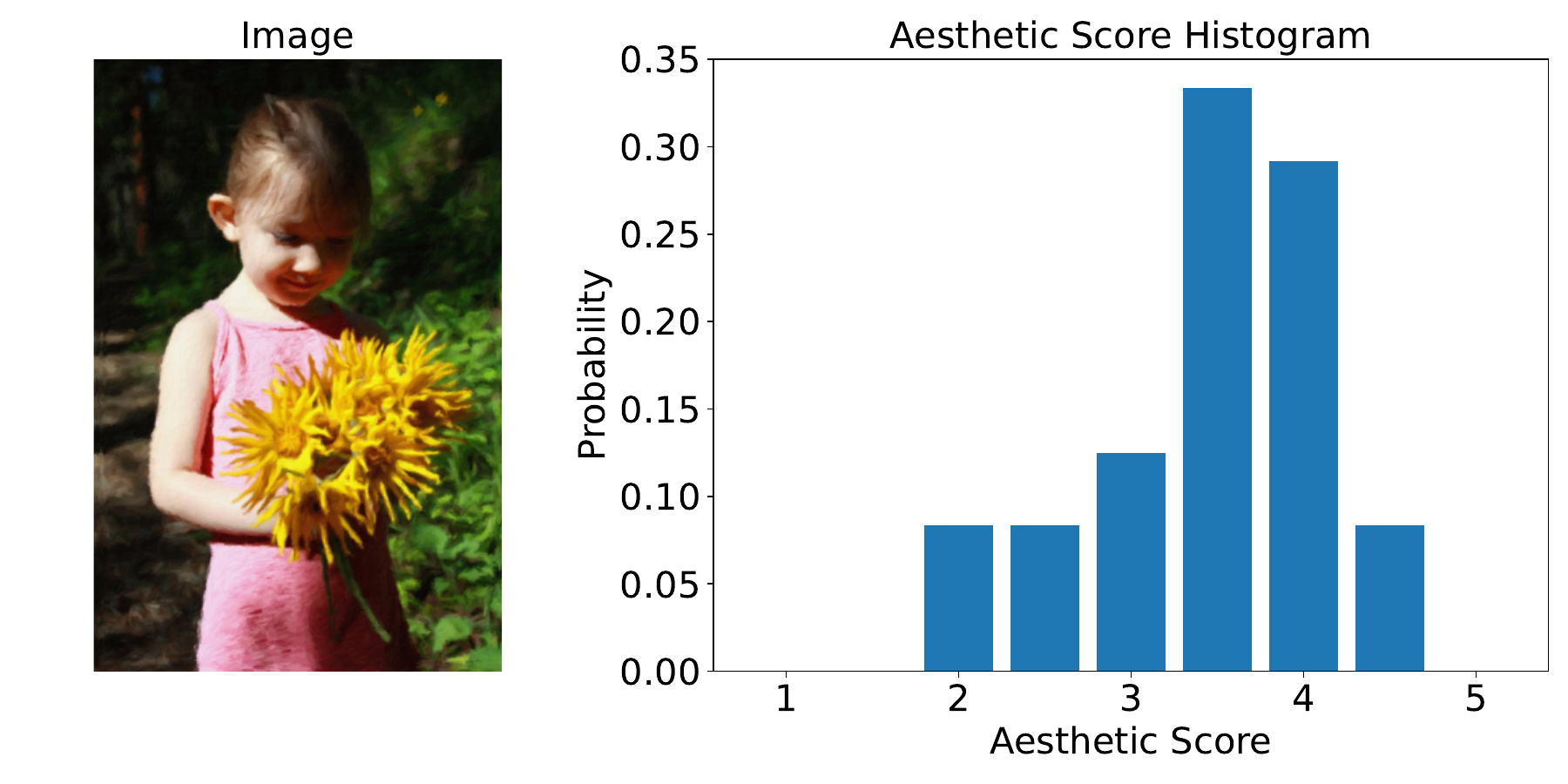}
        \caption{GIAA Data on the PARA dataset~\cite{yang2022personalized}, which consists of an image and a score distribution assembled from individual data. The scores are scaled from 0 to 5 with a spacing of 0.5. The mean score of the score distribution is 3.46.}
        
        \label{fig:dataset_samples:para}
    \end{subfigure}
    \hfill
    \begin{subfigure}[b]{0.48\textwidth}
        \centering
        \includegraphics[width=\linewidth]{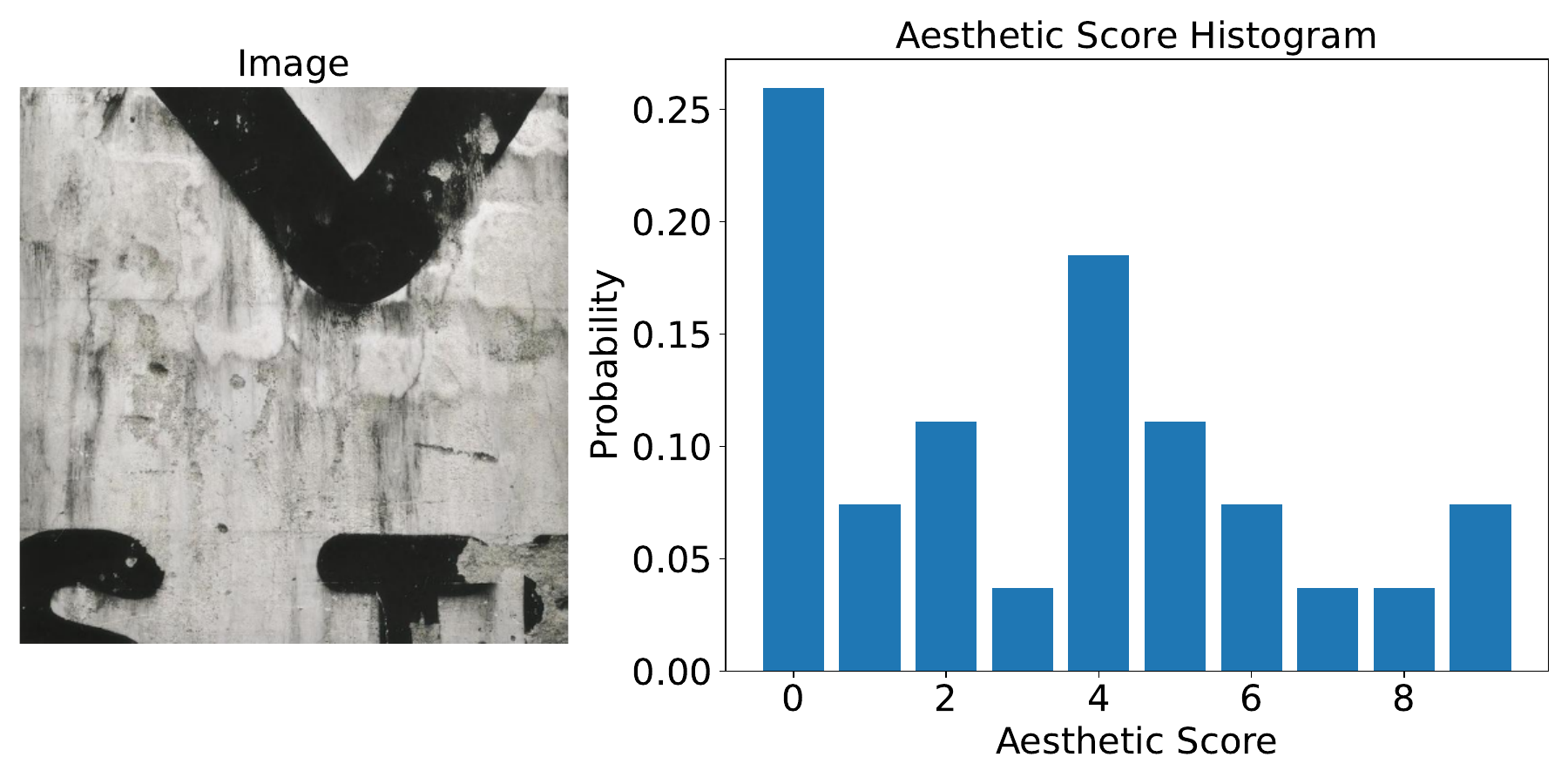}
        \caption{GIAA Data on LAPIS dataset~\cite{maerten2025lapis}, which consists of an image and score distribution assembled from individual data. The scores are scaled from 0-10. The mean score of the score distribution is 3.37.}
        \label{fig:dataset_samples:lapis}
    \end{subfigure}
    \caption{Comparative visualization of GIAA data on the PARA and LAPIS datasets, including score distributions and mean scores.}
    \label{fig:dataset_samples}
\end{figure}

A GIAA example of the PARA dataset~\cite{yang2022personalized} (photos) is shown in Figure~\ref{fig:dataset_samples:para}. The score distribution is assembled from individual scores, as detailed below (user IDs follow personal scores in the dataset):
\begin{samepage}
\begin{itemize}
    \item \textbf{Score 2.0}: Acb3e21, Bcb3b4b
    \item \textbf{Score 2.5}: A64c0cc, B422745
    \item \textbf{Score 3.0}: Abdb7c8, A768d0b, B2eb88a
    \item \textbf{Score 3.5}: A3c6418, A50152a, A09eb7d, A27fab0, Ac9545a, A409131, A5aa4a1, Bab5779
    \item \textbf{Score 4.0}: A7497fb, A697287, A03efa2, A2ce68c, Bbc1ee7, B9042fc, B0fa3ef
    \item \textbf{Score 4.5}: A3f6a35, B6260a3
\end{itemize}
\end{samepage}
where these personal scores are the groundtruth of PIAA data, the same holds for the LAPIS datasets~\cite{maerten2025lapis}. An example of the LAPIS dataset (artworks) is shown in Figure~\ref{fig:dataset_samples:lapis}.

\section{Details of Demography in the LAPIS Dataset}\label{sec:som:demo_detail}
The countries included in the LAPIS dataset are: 'British', 'South African', 'American', 'Portuguese', 'Hungarian', 'Malaysian', 'Belgian', 'Northern Irish', 'Polish', 'Slovenian', 'Spanish', 'Italian', 'Egyptian', 'Scottish', 'Mexican', 'Irish', 'South Korean', 'Greek', 'Czech', 'Brazilian', 'Canadian', 'Indian', 'Ugandan', 'Zimbabwean', 'Dutch', 'Welsh', 'French', 'Finnish', 'German', 'Bangladeshi', 'Lithuanian', 'Australian', 'Tunisian', 'Swiss', 'Romanian', 'Chilean', 'Austrian', 'Nigerien', 'Estonian', 'Bulgarian', 'Turkish', 'Vietnamese', 'Latvian', and 'Malawian'. 
% Besides, the gender categories include male and female as the majority, with non-binary and unknown as minority categories not detailed in the main text.
Besides, the gender categories include male, female, non-binary, and unknown. The latter two, being minority categories, are not mentioned in the main text.
  
\section{Benchmark of PIAA Baselines on PARA}\label{sec:som:mir}
This section discusses the benchmarking of our implementation of the PIAA baselines, including PIAA-MIR~\cite{zhu2022personalized} and PIAA-ICI~\cite{shi2024personalized} models, on the PARA dataset. Note that the original code for these works is not publicly available.
These studies employ a meta-learning scheme for evaluation~\cite{zhu2020personalized}, where a meta-learner is trained on training users and then evaluated on test users, both using few-shot samples. In both stages, the model is fine-tuned on few-shot images and evaluated on the remaining images. In the existing PIAA works on PARA~\cite{yang2023multi,zhu2022personalized,shi2024personalized}, 40 test users were sampled from a total of 438 users and remained fixed throughout the evaluation process. 

Our implemented PIAA-MIR and PIAA-ICI achieve SROCC values of 0.716 and 0.732 with 100-shot fine-tuning, showing comparable performance to the reported values of 0.716~\cite{zhu2022personalized} and 0.739~\cite{shi2024personalized}, respectively.  However, it should be noted that these results may not be strictly comparable, as the test users in this work may differ from those in the original study due to the unavailability of the original test users. Code is available at \url{https://github.com/lwchen6309/aesthetics_transfer_learning}.

\section{Model Architecture and Traits Encoding}\label{sec:som:models}
\begin{figure}[ht]
    \centering
    \includegraphics[width=0.5\linewidth]{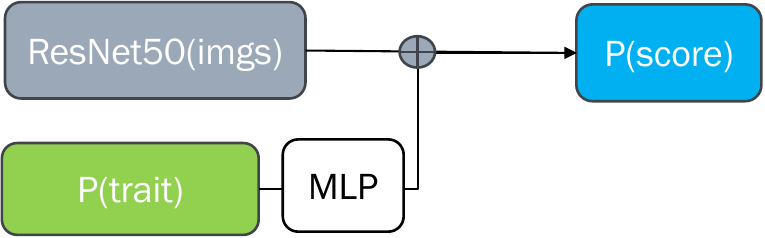}
    \caption{Model architecture of NIMA-trait, which utilize a ResNet-50~\cite{he2016deep} model as the image encoder. The variable \textit{img} represents images, while \textit{P(trait)} and \textit{P(score)} denote the trait distribution and score distribution, respectively.}
    \label{fig:nima_trait}
\end{figure}

\textbf{NIMA-trait}. This section describes the details of the model and data encoding. Figure~\ref{fig:nima_trait} illustrates the architecture of NIMA-trait. The MLP component is a two-layer multilayer perceptron (MLP) with 512 and 10 units in the hidden and output layers, respectively. The input dimension of the MLP corresponds to the trait dimension, which varies based on the trait encoding method used, as discussed below.

\textbf{Trait Encoding}. For trait encoding, the trait dimensions on PARA are 25 and 70, and on LAPIS are 71 and 137, corresponding to conventional encoding and one-hot encoding setups for numeric traits, respectively. The same encoding is also applied to PIAA-MIR~\cite{zhu2022personalized} and PIAA-ICI~\cite{shi2024personalized} in this work.
Specifically, for PARA, the attributes gender, age, educational level, photography experience, and art experience are one-hot encoded into 2, 5, 5, 4, and 4 dimensions, respectively. Combined with the Big-5 personality traits, this results in a total dimension of 25. In the one-hot encoding setup for numeric traits (e.g., Big-5), each Big-5 trait is further one-hot encoded into 10 bins, resulting in a dimension of 50 for all Big-5 traits combined, bringing the total to 70.
For LAPIS, the attributes gender, color blindness, age, educational level, and nationality are one-hot encoded into 4, 2, 5, 5, and 44 dimensions, respectively.  Together with the 11 VAIAK scores, this gives a total dimension of 71. In the one-hot encoding setup for numeric traits (e.g., VAIAK scores), each VAIAK score is further one-hot encoded into 7 bins, yielding a dimension of 77 for all VAIAK scores combined, resulting in a total dimension of 137.

\section{GIAA and PIAA Performance on Overlapped Users}\label{sec:som:seen_benchmark}
The tables in this section report the performance of NIMA, PIAA-ICI, and PIAA-MIR along with their corresponding onehot-encoded models. Table~\ref{table:para_lapis} and Table~\ref{table:para_lapis_plcc} present the model performance in SROCC and PLCC, respectively, when trained with GIAA and PIAA. Similarly, Table~\ref{sgiaa:para_lapis} and Table~\ref{sgiaa:para_lapis_plcc} show the model performance when trained with GIAA and sGIAA.

\begin{table}[]
\centering
\resizebox{.7\columnwidth}{!}{
\begin{tabular}{cccccccc}
\hline
Model & Back- & Train  & \multicolumn{2}{c}{PARA} & \multicolumn{2}{c}{LAPIS} \\ \cline{4-7} 
 & bone & Set & GIAA & PIAA & GIAA & PIAA \\ \hline \hline
\rowcolor{purple!10}
NIMA & & G & \textbf{0.879}~\cite{li2022transductive} & - & \textbf{0.808} & - \\ 
NIMA-trait &  & G & 0.878 & \textcolor{blue}{0.514} & 0.805 & \textcolor{blue}{0.311} \\
PIAA-MIR (Onehot-enc.) &  & G & 0.870 & \textcolor{blue}{0.562} & 0.803 & \textcolor{blue}{0.393} \\
PIAA-ICI (Onehot-enc.) &  & G & 0.873 & \textcolor{blue}{0.578} & 0.798 & \textcolor{blue}{0.389} \\
\rowcolor{cyan!7} 
PIAA-MIR & RN & P & - & 0.717 & - & 0.677 \\ 
\rowcolor{cyan!7} 
PIAA-ICI &  & P & - & 0.730 & - & 0.668 \\ 
NIMA-trait &  & P & \textcolor{red}{0.844} & 0.708 & \textcolor{red}{0.763} & 0.658 \\ %\hline  
PIAA-MIR (Onehot-enc.) &  & P & \textcolor{red}{0.860} & \textbf{0.741} & \textcolor{red}{0.612} & \textbf{0.683} \\ %\hline
PIAA-ICI (Onehot-enc.) &  & P & \textcolor{red}{0.837} & 0.729 & \textcolor{red}{0.803} & 0.660 \\ 
\hline \hline
\rowcolor{purple!10}
NIMA &  & G & \textbf{0.897}~\cite{li2022transductive} & - & \textbf{0.820} & - \\ 
NIMA-trait &  & G & 0.894 & \textcolor{blue}{0.529} & 0.813 & \textcolor{blue}{0.339} \\
PIAA-MIR (Onehot-enc.) &  & G & 0.884 & \textcolor{blue}{0.574} & 0.811 & \textcolor{blue}{0.400} \\
PIAA-ICI (Onehot-enc.) &  & G & 0.886 & \textcolor{blue}{0.572} & 0.813 & \textcolor{blue}{0.398} \\
\rowcolor{cyan!7} 
PIAA-MIR & Swin & P & - & 0.736 & - & 0.686 \\ 
\rowcolor{cyan!7} 
PIAA-ICI &  & P & - & 0.737 & - & 0.677 \\ 
NIMA-trait &  & P & \textcolor{red}{0.856} & 0.720 & \textcolor{red}{0.806} & 0.654 \\
PIAA-MIR (Onehot-enc.) &  & P & \textcolor{red}{0.870} & \textbf{0.749} & \textcolor{red}{0.752} & \textbf{0.693} \\
PIAA-ICI (Onehot-enc.) &  & P & \textcolor{red}{0.885} & 0.745 & \textcolor{red}{0.775} & 0.670 \\ 
\hline \hline
\rowcolor{purple!10}
NIMA &  & G & 0.890 & - & 0.825  & - \\ 
NIMA-trait &  & G & \textbf{0.897} & \textcolor{blue}{0.544} & \textbf{0.830} & \textcolor{blue}{0.338} \\
PIAA-MIR (Onehot-enc.) &  & G & 0.888 & \textcolor{blue}{0.555} & 0.820 & \textcolor{blue}{0.384} \\

PIAA-ICI (Onehot-enc.) &  & G & 0.891 & \textcolor{blue}{0.576} & 0.814 & \textcolor{blue}{0.387} \\
\rowcolor{cyan!7} 
PIAA-MIR & ViT & P & - & 0.727 & - & 0.683 \\ 
\rowcolor{cyan!7} 
PIAA-ICI &  & P & - & 0.732 & - & 0.665 \\ 
NIMA-trait &  & P & \textcolor{red}{0.848} & 0.716 & \textcolor{red}{0.624} & 0.656 \\ 
PIAA-MIR (Onehot-enc.) &  & P & \textcolor{red}{0.867} & \textbf{0.743} & \textcolor{red}{0.709} & \textbf{0.688} \\ 
PIAA-ICI (Onehot-enc.) &  & P & \textcolor{red}{0.871} & 0.737 & \textcolor{red}{0.722} & 0.662 \\ 
\hline
\end{tabular}
}
\vspace{0.4cm}
\caption{\textbf{SROCC of IAA models trained on GIAA and PIAA using the PARA and LAPIS datasets.} Abbreviations for the training sets: G = GIAA and P = PIAA. The models RN, Swin, and ViT refer to ResNet50, Swin-tiny, and ViT-small, respectively. The zero-shot performance of PIAA (i.e., trained on GIAA and tested on PIAA) is shown in \textcolor{blue}{blue}, while the zero-shot performance of GIAA (i.e., trained on PIAA and tested on GIAA) is shown in \textcolor{red}{red}. Our models achieve comparable performance to the \colorbox{purple!10}{GIAA baseline (\ie, NIMA)} and \colorbox{cyan!7}{PIAA baselines (\ie, PIAA-MIR and PIAA-ICI)}.
}
\label{table:para_lapis}
\end{table}

\begin{table}[]
\centering
\resizebox{.7\columnwidth}{!}{
\begin{tabular}{cccccccc}
\hline
Model & Back- & Train  & \multicolumn{2}{c}{PARA} & \multicolumn{2}{c}{LAPIS} \\ \cline{4-7} 
 & bone & Set & GIAA & PIAA & GIAA & PIAA \\ \hline \hline
\rowcolor{purple!10}
NIMA &  & G & \textbf{0.921}~\cite{li2022transductive} & - & \textbf{0.812} & - \\ 
NIMA-trait & & G & 0.915 & \textcolor{blue}{0.563} & 0.805 & \textcolor{blue}{0.316} \\
PIAA-MIR (Onehot-enc.) &  & G & 0.909 & \textcolor{blue}{0.623} & 0.806 & \textcolor{blue}{0.395} \\
PIAA-ICI (Onehot-enc.) &  & G & 0.911 & \textcolor{blue}{0.635} & 0.804 & \textcolor{blue}{0.392} \\
\rowcolor{cyan!7} 
PIAA-MIR & RN & P & - & 0.748 & - & 0.687 \\ 
\rowcolor{cyan!7} 
PIAA-ICI &  & P & - & 0.760 & - & 0.678 \\ 
NIMA-trait &  & P & \textcolor{red}{0.870} & 0.727 & \textcolor{red}{0.764} & 0.666 \\
PIAA-MIR (Onehot-enc.) &  & P & \textcolor{red}{0.896} & \textbf{0.770} & \textcolor{red}{0.639} & \textbf{0.692} \\
PIAA-ICI (Onehot-enc.) &  & P & \textcolor{red}{0.891} & 0.760 & \textcolor{red}{0.807} & 0.670 \\
\hline \hline
\rowcolor{purple!10}
NIMA &  & G & \textbf{0.933}~\cite{li2022transductive} & - & \textbf{0.825}  & - \\ 
NIMA-trait & & G & 0.928 & \textcolor{blue}{0.579} & 0.818 & \textcolor{blue}{0.342} \\
PIAA-MIR (Onehot-enc.) &  & G & 0.922 & \textcolor{blue}{0.631} & 0.815 & \textcolor{blue}{0.400} \\
PIAA-ICI (Onehot-enc.) &  & G & 0.923 & \textcolor{blue}{0.630} & 0.817 & \textcolor{blue}{0.399} \\
\rowcolor{cyan!7}
PIAA-MIR & Swin & P & - & 0.765 & - & 0.696 \\  
\rowcolor{cyan!7}
PIAA-ICI &  & P & - & 0.767 & - & 0.687 \\ 
NIMA-trait &  & P & \textcolor{red}{0.885} & 0.737 & \textcolor{red}{0.791} & 0.662 \\
PIAA-MIR (Onehot-enc.) &  & P & \textcolor{red}{0.906} & \textbf{0.777} & \textcolor{red}{0.757} & \textbf{0.701} \\
PIAA-ICI (Onehot-enc.) &  & P & \textcolor{red}{0.923} & 0.775 & \textcolor{red}{0.788} & 0.680 \\
\hline \hline
\rowcolor{purple!10}
NIMA &  & G & 0.927  & - & 0.828  & - \\ 
NIMA-trait & & G & \textbf{0.931} & \textcolor{blue}{0.602} & \textbf{0.834} & \textcolor{blue}{0.341} \\
PIAA-MIR (Onehot-enc.) &  & G & 0.925 & \textcolor{blue}{0.617} & 0.824 & \textcolor{blue}{0.382} \\
PIAA-ICI (Onehot-enc.) &  & G & 0.927 & \textcolor{blue}{0.638} & 0.819 & \textcolor{blue}{0.390} \\
\rowcolor{cyan!7}
PIAA-MIR & ViT & P & - & 0.757 & - & 0.692  \\ 
\rowcolor{cyan!7}
PIAA-ICI &  & P & - & 0.761 & - & 0.674  \\ 
NIMA-trait &  & P & \textcolor{red}{0.884} & 0.734 & \textcolor{red}{0.635} & 0.664 \\
PIAA-MIR (Onehot-enc.) &  & P & \textcolor{red}{0.906} & \textbf{0.771} & \textcolor{red}{0.720} & \textbf{0.697} \\
PIAA-ICI (Onehot-enc.) &  & P & \textcolor{red}{0.911} & 0.766 & \textcolor{red}{0.732} & 0.672 \\
\hline
\end{tabular}
}
\vspace{0.4cm}
\caption{\textbf{PLCC of IAA models trained on GIAA and PIAA using the PARA and LAPIS Datasets.} Abbreviations for the training sets: G = GIAA and P = PIAA. The models RN, Swin, and ViT refer to ResNet50, Swin-tiny, and ViT-small, respectively. The zero-shot performance of PIAA (i.e., trained on GIAA and tested on PIAA) is shown in \textcolor{blue}{blue}, while the zero-shot performance of GIAA (i.e., trained on PIAA and tested on GIAA) is shown in \textcolor{red}{red}. Our models achieve comparable performance to the \colorbox{purple!10}{GIAA baseline (\ie, NIMA)} and \colorbox{cyan!7}{PIAA baselines (\ie PIAA-MIR and PIAA-ICI)}.
}
\label{table:para_lapis_plcc}
\end{table}

\begin{table}[]
\centering
\resizebox{0.7\columnwidth}{!}{
\begin{tabular}{cccccccc}
\hline
Model & Back- & Train  & \multicolumn{2}{c}{PARA} & \multicolumn{2}{c}{LAPIS} \\ \cline{4-7} 
 & bone & Set & GIAA & PIAA & GIAA & PIAA \\ \hline \hline
\multirow{2}{*}{NIMA-trait} & \multirow{6}{*}{RN} & G & 0.878 & \textcolor{blue}{0.514} & 0.805 & \textcolor{blue}{0.311} \\
 &  & sG & 0.875 & \textcolor{blue}{0.650 \textbf{(+27\%)}} & 0.802 & \textcolor{blue}{0.447 \textbf{(+44\%)}} \\
PIAA-MIR &  & G & 0.870 & \textcolor{blue}{0.562} & 0.803 & \textcolor{blue}{0.393} \\
(Onehot-enc.) &  & sG & 0.868 & \textcolor{blue}{0.635 (+13\%)} & 0.803 & \textcolor{blue}{0.423 (+8\%)} \\ 
PIAA-ICI &  & G & 0.873 & \textcolor{blue}{0.578} & 0.798 & \textcolor{blue}{0.389} \\
(Onehot-enc.) & & sG & 0.870 & \textcolor{blue}{0.596 (+3\%)} & 0.805 & \textcolor{blue}{0.405 (+4\%)} \\
\hline \hline
\multirow{2}{*}{NIMA-trait} & \multirow{6}{*}{Swin} & G & 0.894 & \textcolor{blue}{0.529} & 0.813 & \textcolor{blue}{0.339} \\
  &  & sG & 0.893 & \textcolor{blue}{0.660 \textbf{(+25\%)}} & 0.814 & \textcolor{blue}{0.454 \textbf{(+34\%)}} \\
PIAA-MIR &  & G & 0.884 & \textcolor{blue}{0.574} & 0.811 & \textcolor{blue}{0.400} \\
(Onehot-enc.) &  & sG & 0.881 & \textcolor{blue}{0.648 (+13\%)} & 0.811 & \textcolor{blue}{0.439 (+10\%)} \\
PIAA-ICI &  & G & 0.886 & \textcolor{blue}{0.572} & 0.813 & \textcolor{blue}{0.398} \\
(Onehot-enc.) &  & sG & 0.887 & \textcolor{blue}{0.617 (+8\%)} & 0.809 & \textcolor{blue}{0.414 (+4\%)} \\ 
\hline \hline
\multirow{2}{*}{NIMA-trait} & \multirow{6}{*}{ViT} & G & 0.897 & \textcolor{blue}{0.544} & 0.830 & \textcolor{blue}{0.338} \\
 &  & sG & 0.894 & \textcolor{blue}{0.658 \textbf{(+21\%)}} & 0.827 & \textcolor{blue}{0.450 \textbf{(+33\%)}} \\ 
PIAA-MIR &  & G & 0.888 & \textcolor{blue}{0.555} & 0.820 & \textcolor{blue}{0.384} \\
(Onehot-enc.) &  & sG & 0.887 & \textcolor{blue}{0.658 (+18\%)} & 0.819 & \textcolor{blue}{0.445 (+16\%)} \\ 
PIAA-ICI &  & G & 0.891 & \textcolor{blue}{0.576} & 0.814 & \textcolor{blue}{0.387} \\
(Onehot-enc.) &  & sG & 0.887 & \textcolor{blue}{0.619 (+7\%)} & 0.821 & \textcolor{blue}{0.417 (+8\%)} \\
\hline
\end{tabular}
}
\vspace{0.4cm}
\caption{\textbf{SROCC of IAA models trained on GIAA and sGIAA using the PARA and LAPIS datasets.} Train set abbreviations: G = GIAA and sG = sGIAA. The format follows that of Table~\ref{table:para_lapis}. The increase in model performance when trained on sGIAA compared to GIAA is shown in parentheses.}
\label{sgiaa:para_lapis}
\end{table}

\begin{table}[]
\centering
\resizebox{.7\columnwidth}{!}{
\begin{tabular}{cccccccc}
\hline
Model & Back- & Train  & \multicolumn{2}{c}{PARA} & \multicolumn{2}{c}{LAPIS} \\ \cline{4-7} 
 & bone & Set & GIAA & PIAA & GIAA & PIAA \\ \hline \hline
\multirow{2}{*}{NIMA-trait} & \multirow{6}{*}{RN50} & G & 0.915 & \textcolor{blue}{0.563} & 0.805 & \textcolor{blue}{0.316} \\
 &  & sG & 0.916 & \textcolor{blue}{0.688 \textbf{(+22\%)}} & 0.807 & \textcolor{blue}{0.454 \textbf{(+44\%)}} \\
PIAA-MIR &  & G & 0.909 & \textcolor{blue}{0.623} & 0.806 & \textcolor{blue}{0.395} \\
(Onehot-enc.) &  & sG & 0.908 & \textcolor{blue}{0.675 (+8\%)} & 0.809 & \textcolor{blue}{0.427 (+8\%)} \\
PIAA-ICI &  & G & 0.911 & \textcolor{blue}{0.635} & 0.804 & \textcolor{blue}{0.392} \\
(Onehot-enc.) & & sG & 0.910 & \textcolor{blue}{0.650 (+2\%)} & 0.810 & \textcolor{blue}{0.409 (+4\%)} \\
\hline \hline
\multirow{2}{*}{NIMA-trait} & \multirow{6}{*}{Swin-t} & G & 0.928 & \textcolor{blue}{0.579} & 0.818 & \textcolor{blue}{0.342} \\
  &  & sG & 0.928 & \textcolor{blue}{0.696 \textbf{(+20\%)}} & 0.815 & \textcolor{blue}{0.460 \textbf{(+34\%)}} \\
PIAA-MIR &  & G & 0.922 & \textcolor{blue}{0.631} & 0.815 & \textcolor{blue}{0.400} \\
(Onehot-enc.) &  & sG & 0.920 & \textcolor{blue}{0.688 (+9\%)} & 0.816 & \textcolor{blue}{0.443 (+11\%)} \\
PIAA-ICI &  & G & 0.923 & \textcolor{blue}{0.630} & 0.817 & \textcolor{blue}{0.399} \\
(Onehot-enc.) &  & sG & 0.925 & \textcolor{blue}{0.667 (+6\%)} & 0.815 & \textcolor{blue}{0.417 (+4\%)} \\
\hline \hline
\multirow{2}{*}{NIMA-trait} & \multirow{6}{*}{ViT-s} & G & 0.931 & \textcolor{blue}{0.602} & 0.834 & \textcolor{blue}{0.341} \\
 &  & sG & 0.927 & \textcolor{blue}{0.695 \textbf{(+15\%)}} & 0.831 & \textcolor{blue}{0.455 \textbf{(+34\%)}} \\
PIAA-MIR &  & G & 0.925 & \textcolor{blue}{0.617} & 0.824 & \textcolor{blue}{0.382} \\
(Onehot-enc.) &  & sG & 0.924 & \textcolor{blue}{0.695 (+13\%)} & 0.821 & \textcolor{blue}{0.450 (+18\%)} \\
PIAA-ICI &  & G & 0.927 & \textcolor{blue}{0.638} & 0.819 & \textcolor{blue}{0.390} \\
(Onehot-enc.) &  & sG & 0.923 & \textcolor{blue}{0.667 (+5\%)} & 0.826 & \textcolor{blue}{0.419 (+8\%)} \\
\hline
\end{tabular}
}
\vspace{0.4cm}
\caption{\textbf{PLCC of IAA models trained on GIAA and sGIAA using the PARA and LAPIS Datasets.} Train set abbreviations: G = GIAA and sG = sGIAA. The format follows that of Table~\ref{table:para_lapis_plcc}. The increase in model performance when trained on sGIAA compared to GIAA is shown in parentheses.}
\label{sgiaa:para_lapis_plcc}
\end{table}

\section{Individual Subjectivity of Image Aesthetic}
This section examines the variations in score distribution across users from distinct demographic groups. We utilize the Gini index to illustrate how the distribution of aesthetic scores varies across different demographic splits. A lower Gini index indicates a more effective demographic split, enabling clearer distinctions in the distribution of aesthetic scores among user groups. 
% We perform splits based on age, educational level, and gender for both the PARA and LAPIS datasets. Additionally, we consider splits based on art experience and photography experience in the PARA dataset, and VAIAK traits in the LAPIS dataset.
The results are presented in Table~\ref{table:gini}; the Gini index for VAIAK1-7 in the LAPIS dataset is excluded due to its consistently high and nearly constant value ($0.769 \pm 0.009$), suggesting limited relevance to aesthetic variation across users\footnote{The Gini index values for VAIAK1-7 are 0.764, 0.776, 0.779, 0.764, 0.777, 0.771, and 0.750, respectively}.
It is evident that in both datasets, the Gini index is high for gender, suggesting that it does not significantly impact aesthetic preferences. Conversely, a lower Gini index is observed for educational level and photography experience in the PARA dataset, while in the LAPIS dataset, lower indices are noted for educational level, age, and specific VAIAK traits (particularly 2VAIAK1 and 2VAIAK4).

\begin{table}[h]
\centering
\caption{Gini index values for different demographic splits on the PARA and LAPIS datasets. A low Gini index indicates a more effective demographic split for distinguishing aesthetic preferences.}
\label{table:gini}

\parbox{.4\linewidth}{
\centering
\resizebox{\linewidth}{!}{
\begin{tabular}{c|c}
Trait & Gini Index ($\downarrow$) \\ \hline \hline
Age & 0.553 \\
ArtExperience & 0.550 \\
EducationalLevel & \textbf{0.461} \\
Gender & 0.677 \\
PhotographyExperience & 0.495 \\
\end{tabular}
}
\subcaption{PARA dataset}
\vspace{-0.8cm}
\label{table:gini:para}
}
\parbox{.36\linewidth}{
\centering
\resizebox{\linewidth}{!}{
\begin{tabular}{c|c}
Trait & Gini Index  ($\downarrow$) \\ \hline \hline
Age & 0.489 \\
EducationalLevel & \textbf{0.423} \\
Gender & 0.770 \\
2VAIAK1 & 0.466 \\
2VAIAK2 & 0.617 \\
2VAIAK3 & 0.779 \\
2VAIAK4 & 0.456 \\
\end{tabular}
}
\subcaption{LAPIS dataset.}
\vspace{-0.8cm}
\label{table:gini:lapis}
}
\end{table}

\section{GIAA and PIAA Performance on Unseen Users}\label{sec:som:benchmark}
\begin{figure}[htbp]
    \centering
    \begin{subfigure}[b]{0.48\linewidth}
        \centering
        \includegraphics[width=1.\linewidth]{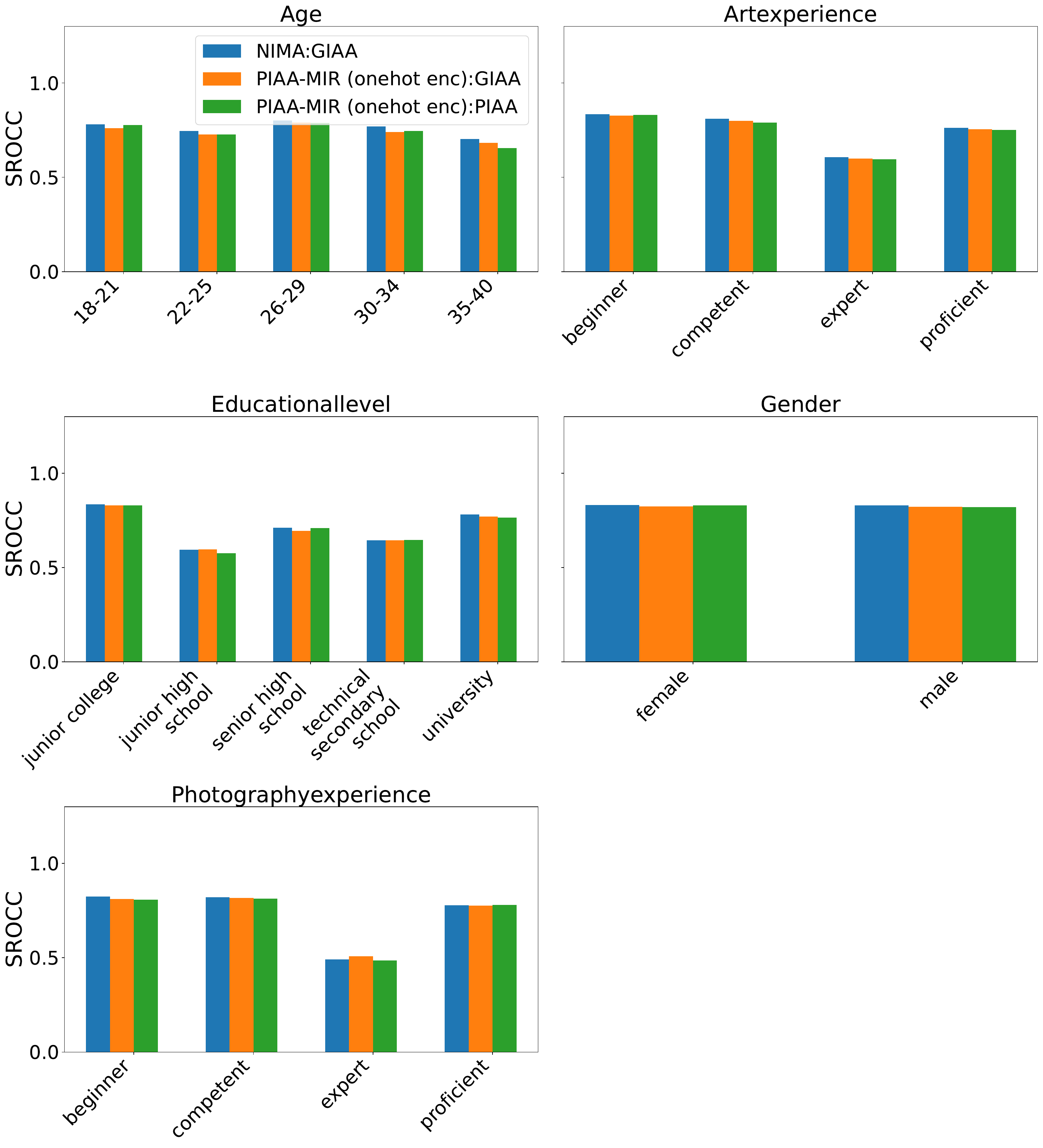}
        \caption{GIAA SROCC.}
        \label{fig:srocc:para_giaa}
    \end{subfigure}
    \hfill
    \begin{subfigure}[b]{0.48\linewidth}
        \centering
        \includegraphics[width=1.\linewidth]{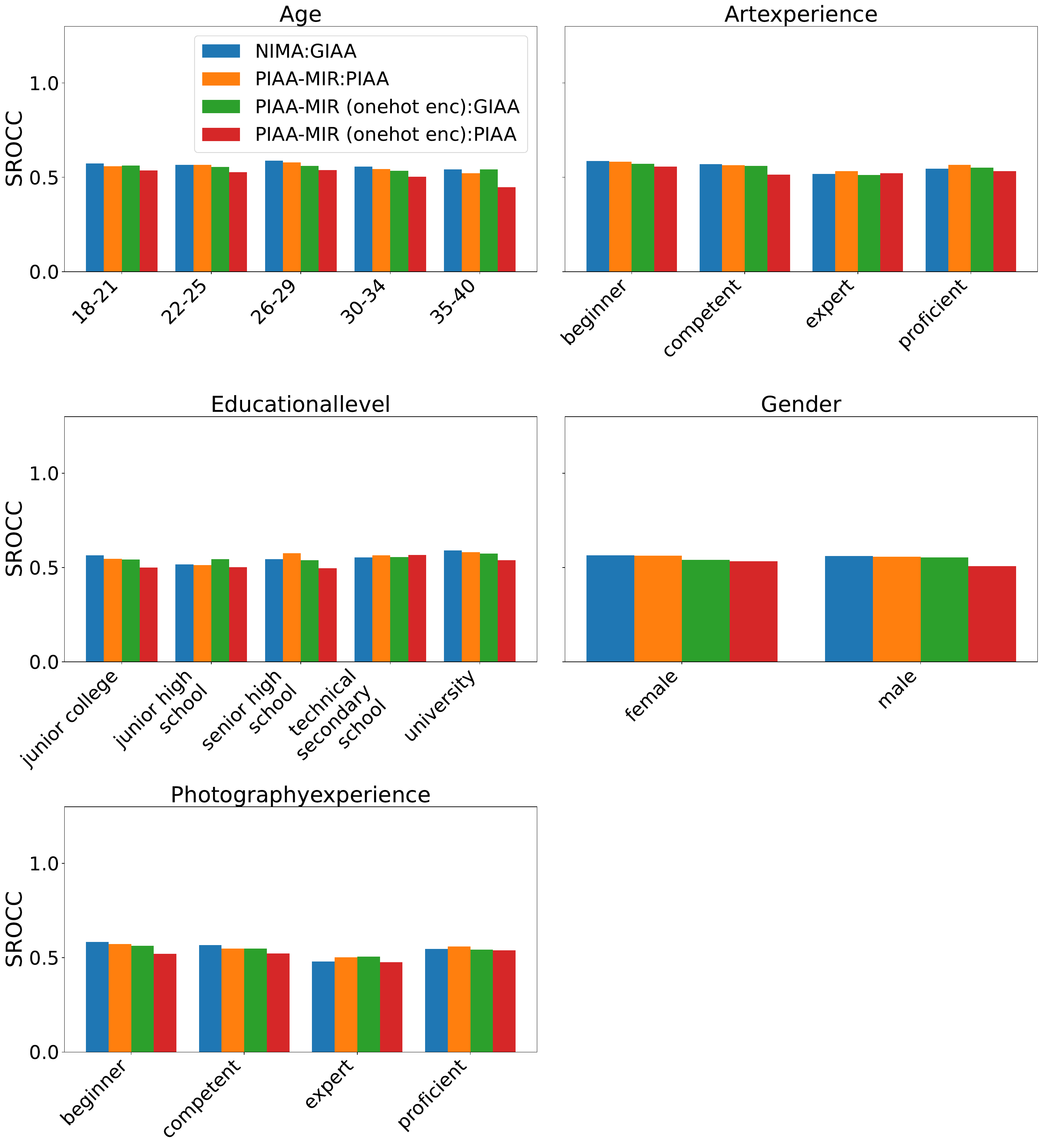}
        \caption{PIAA SROCC.}
        \label{fig:srocc:para_piaa}
    \end{subfigure}
    \caption{SROCC performance of models trained on disjoint users from the PARA dataset.}
    \label{fig:srocc:para}
\end{figure}

\begin{figure}[h]
    \centering
    \begin{subfigure}[b]{0.48\linewidth}
        \centering
        \includegraphics[width=\linewidth]{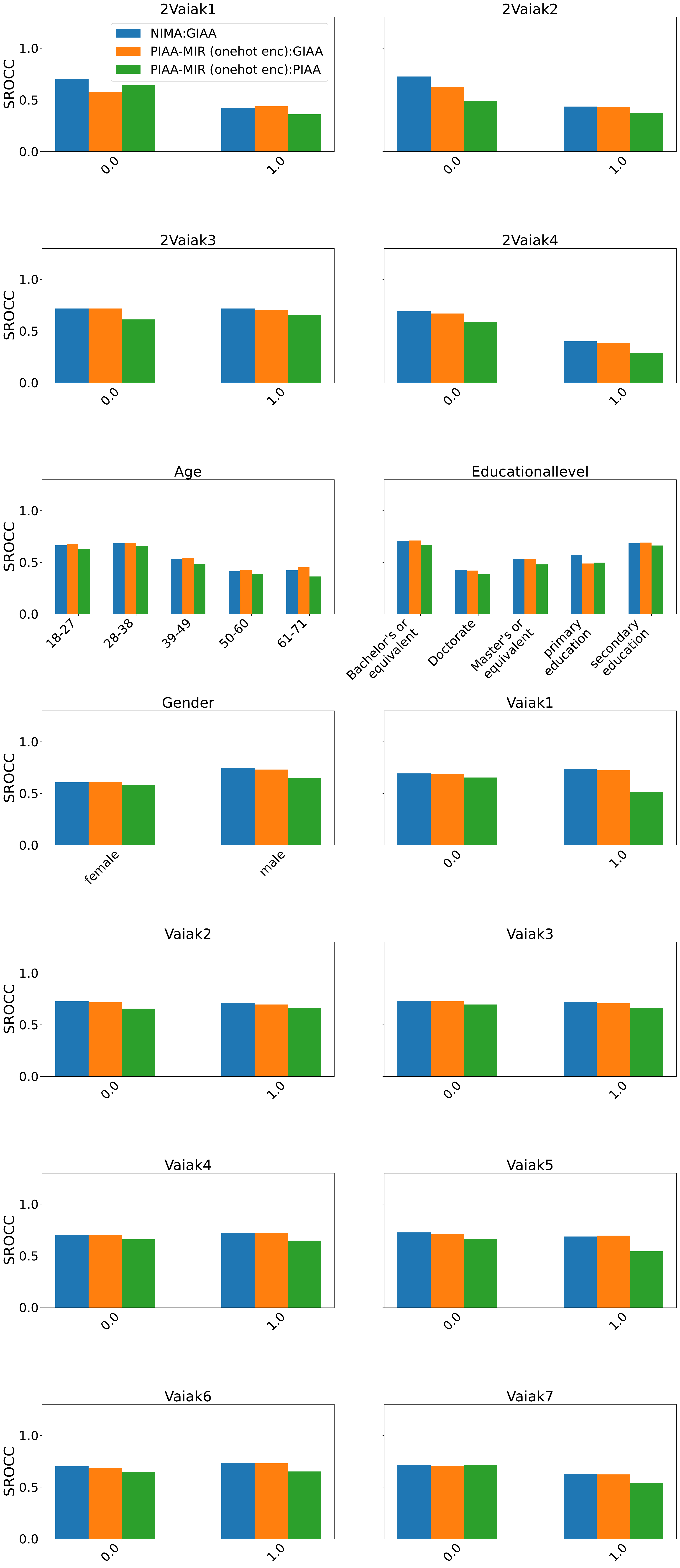}
        \caption{GIAA SROCC.}
        \label{fig:srocc:lapis_giaa}
    \end{subfigure}
    \hfill
    \begin{subfigure}[b]{0.48\linewidth}
        \centering
        \includegraphics[width=\linewidth]{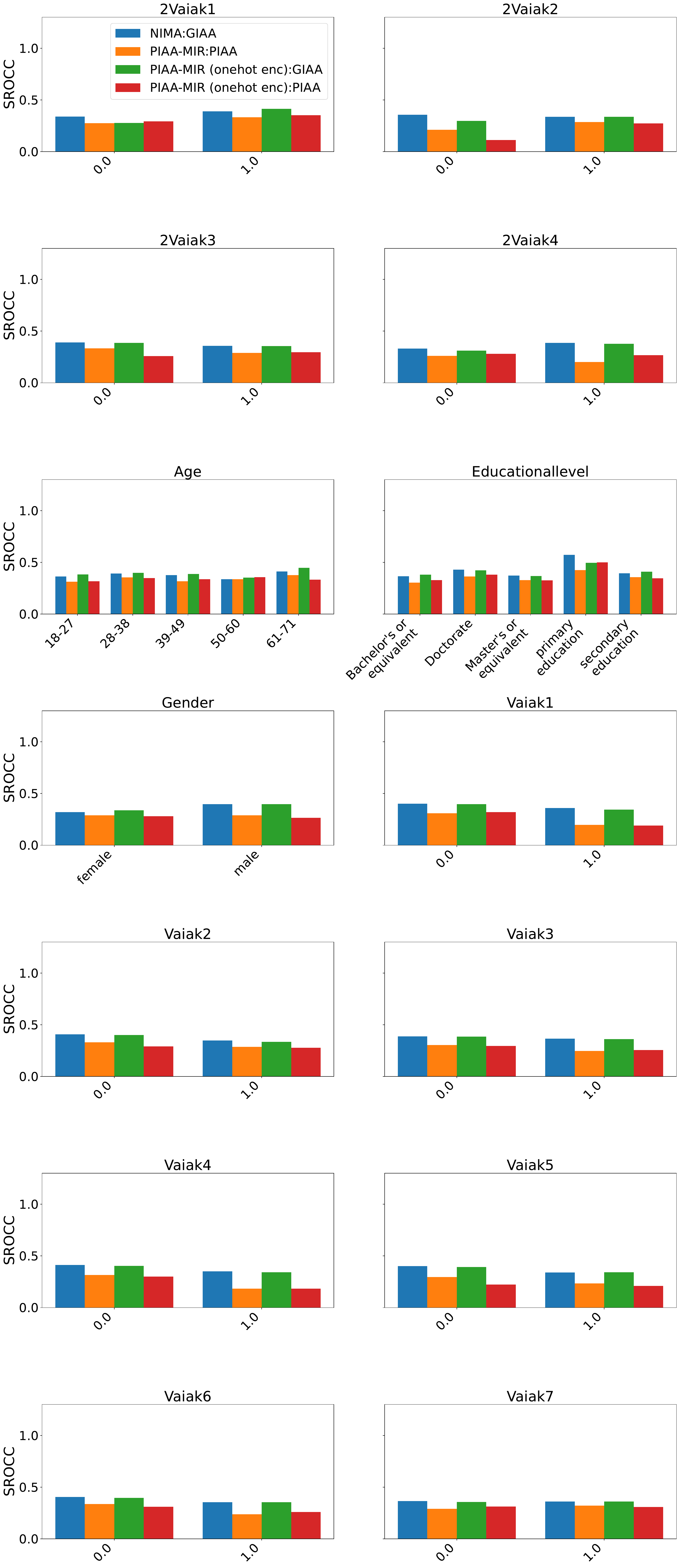}
        \caption{PIAA SROCC.}
        \label{fig:srocc:lapis_piaa}
    \end{subfigure}
    \caption{SROCC performance of models trained on disjoint users from the LAPIS dataset.}
    \label{fig:srocc:lapis}
\end{figure}

Using ResNet 50 as backbone, we evaluate the performance of NIMA, PIAA-MIR, and PIAA-MIR (Onehot-enc.) across different demographic splits to explore the model generalization to unseen users. 
For the GIAA performance, NIMA, PIAA-MIR (Onehot-enc.) trained on both GIAA and PIAA are evaluated. Their performance is depicted in Figure~\ref{fig:srocc:para_giaa} and Figure~\ref{fig:srocc:lapis_giaa} on PARA and LAPIS datasets, respectively. 
We observe a trend similar to the EMD results shown in Figure~4 of the main text. For the PARA dataset, the SROCC values are low for experts in both photography and art, as well as for users with only junior high school education, highlighting their distinct aesthetic preferences for photos. For the LAPIS dataset, the SROCC values are low for older users, users with either doctoral or primary-level education, and those with extensive art experience (especially for 2VAIAK1-4), indicating their distinct aesthetic preferences for artworks.

For the PIAA performance, PIAA-MIR is additionally included. The model performances are depicted in Figure~\ref{fig:srocc:para_piaa} and Figure~\ref{fig:srocc:lapis_piaa} show their performance on PARA and LAPIS datasets, respectively.
Compared to the GIAA evaluation scheme, we observe similar PIAA performance across various demographic splits, suggesting that model inference under the PIAA evaluation scheme may be more robust for unseen users. Although the underlying rationale is not yet fully clear, it is possible that individual subjectivity is crucial for the model’s generalization to unseen users, while such subjectivity is suppressed by averaging individual data in GIAA.
Moreover, we observe similar performance between the PIAA-MIR (Onehot-enc.) models trained on GIAA and PIAA, suggesting that the interpolation and extrapolation discussed in our transfer learning theory do not hold here. The reason is that our theory addresses the difference between individual distributions and averaged distributions within a single convex hull (either in trait space or score space), where all users are accessible to the model during the training phase. However, with an unseen user setup, this configuration splits the convex hull into two parts: one for training users and one for test users, thus rendering the discussion of interpolation and extrapolation inapplicable. We aim to further explore our theory under these conditions in future work.

\clearpage
\bibliography{egbib}